# ReMoDEx: A Local-to-Global Relevance-Based Model Decision Explainability Framework for large-Scale Image Datasets

Abhay Kumar Pathak[a], Mrityunjay Chaubey[b], Manjari Gupta[a,*]

[a] *Department of Computer Science, Institute of Science, Banaras Hindu University, Varanasi, India*

[b] *University of Petroleum and Energy Studies, Dehradun, India*

*** Corresponding author**: Manjari Gupta *( manjari@bhu.ac.in )*

Abhay Kumar Pathak *( pathakabhay@bhu.ac.in )*

Mrityunjay Chaubey ( *mrityunjay.chaubey@ddn.upes.ac.in*)

**Abstract**: Deep learning image classifiers achieve strong predictive performance yet remain opaque in how decisions are formed. A model may predict correctly while relying on irrelevant cues, shortcut associations, peripheral structures, or device level artifacts instead of task relevant regions. On large scale datasets this opacity is especially problematic, since inspecting heatmaps one sample at a time cannot scale to thousands of predictions. We propose Relevance Based Model Decision Explainability (ReMoDEx), a framework for systematic, dataset scale assessment of model decision behaviour in image classification. ReMoDEx defines a stepwise pipeline: model inference, target class selection, relevance map generation, heatmap standardisation, similarity based grouping of patterns, cluster level interpretation, and spatial relevance assessment. Local methods GradCAM++, Integrated Gradients, Occlusion Sensitivity, and Layerwise Relevance Propagation are each combined independently with a single global module that summarises an entire set of relevance maps into a few decision strategy clusters, replacing sample by sample inspection with an automatic, scalable summary. To demonstrate ReMoDEx, we applied it to a VGG16 based classifier distinguishing COVID-19, Normal, Lung Opacity, and Viral Pneumonia. The classifier showed stable performance (86.27% test accuracy, 0.9624 test AUC). However, each explainer combined with the global module consistently produced two recurring strategies: central thoracic region decisions and border/corner sensitive decisions, indicating possible shortcut learning that conventional metrics could not reveal. Masked image validation confirmed that model confidence and predicted class changed when central or peripheral regions were occluded. ReMoDEx thus provides a scalable relevance based decision assessment framework and an essential complement to accuracy based evaluation.

# 1. Introduction

Deep learning image classifiers learn hierarchical visual representations directly from image data, allowing them to perform classification without depending on manually designed feature descriptors (LeCun et al., 2015; Litjens et al., 2017). This representation-learning ability has made deep neural networks highly effective for image classification, but the model output usually provides only the predicted class or a probability score. The predicted label does not directly reveal which image regions, visual structures, or spatial patterns contributed to the decision. Consequently, a model can appear accurate at the output level while its internal decision process remains unclear. This lack of transparency matters because image classifiers may learn visual correlations that are useful for separating classes in the training data but are not necessarily related to the intended object, pathology, or task-specific evidence. Such correlations may arise from background regions, image borders, acquisition settings, preprocessing differences, text markers, or other dataset-dependent visual patterns. In these situations the prediction may be correct while the visual basis of the decision is not meaningful for the intended task. Conventional classification metrics—accuracy, precision, recall, F1-score, and AUC measure predictive agreement, but they do not determine whether the model used relevant image evidence during decision-making. Decision transparency is therefore not only a matter of explaining individual predictions, but also of examining whether the learned behaviour is consistent with meaningful, task-relevant visual reasoning. This issue becomes more critical when the model repeatedly depends on unintended cues, because such dependence can produce apparently strong performance while masking shortcut-driven behaviour.

This concern becomes more specific when a model repeatedly relies on unintended visual cues rather than task-relevant image evidence. This behaviour is commonly described as shortcut learning (Geirhos et al., 2020), in which the classifier learns an easier association present in the dataset instead of the intended visual basis of classification. In image classification, shortcut learning may occur when class labels become associated with non-target patterns such as image borders, corner regions, background structures, embedded labels, contrast differences, cropping style, scanner-specific patterns, or preprocessing artifacts. Such cues may help separate classes during internal training and testing, but they do not necessarily represent meaningful evidence for the actual task. The Clever Hans effect (Lapuschkin et al., 2019) is an extreme form of this problem: the model appears to solve the task correctly, but its decision is based on cues that are only predictive within the development dataset. A correct prediction therefore cannot automatically be interpreted as a meaningful decision, because the same label may be produced from clinically or semantically irrelevant regions. The risk is greatest when the shortcut cue is consistently present in the training and internal testing data, because conventional evaluation may still report acceptable performance. When the model is later applied to images from a different source, device, acquisition protocol, or preprocessing pipeline, the shortcut cue may change or disappear, leading to unstable generalisation (Zech et al., 2018). Shortcut learning and Clever Hans-type behaviour thus cannot be reliably identified from the predicted class or performance score alone; they require direct examination of the spatial evidence used during prediction.

Because shortcut learning can still produce correct predictions, conventional performance metrics alone cannot verify whether a model has learned the intended decision strategy. Accuracy, precision, recall, F1-score, and AUC measure agreement between predicted and true labels, but they do not indicate which image regions contributed to those predictions. Two models with similar classification performance may therefore rely on different visual evidence: one on task-relevant regions and another on background patterns, border structures, acquisition artifacts, or other shortcut-sensitive cues. This limitation is amplified when test data are drawn from the same or a similar distribution as the training data, because shortcut cues may remain present in both sets. In such conditions a model can show stable internal performance while still depending on evidence that does not generalise externally. Accuracy-based evaluation can confirm that a prediction is correct, but not that the basis of the prediction is meaningful. For this reason model evaluation should include decision-level analysis of the spatial relevance pattern supporting each prediction, so that it can be determined whether correct predictions are consistently supported by relevant regions or whether some are influenced by shortcut-sensitive structures.

## 1.1 Need for local-to-global relevance-based explainability

The limitations of accuracy-based evaluation indicate that model assessment must consider the visual evidence used during prediction, not only the final predicted class. Local explainability methods address this requirement by producing prediction-specific relevance maps that indicate which regions contributed to the output. For convolutional classifiers, Grad-CAM and Grad-CAM++ generate class-discriminative heatmaps from convolutional feature maps (Selvaraju et al., 2017; Chattopadhay et al., 2018), allowing visual inspection of the regions that influenced a specific prediction. Layer-wise Relevance Propagation provides another form of relevance estimation by propagating the prediction score backward through the network and assigning relevance to input regions (Bach et al., 2015). Perturbation-based approaches such as occlusion sensitivity examine model dependence on regions by measuring how prediction confidence changes when selected regions are masked (Zeiler & Fergus, 2014). Complementary attribution and interpretability methods, together with broad surveys of explainable AI, situate these tools within a wider methodological landscape (Sundararajan et al., 2017; Ribeiro et al., 2016; Lundberg & Lee, 2017; Montavon et al., 2018; Samek et al., 2021; Arrieta et al., 2020; Tjoa & Guan, 2021; Amann et al., 2020). These methods are useful for individual predictions, but a single-sample explanation cannot fully describe the decision behaviour of the complete trained model. A model may show meaningful relevance in some samples and shortcut-sensitive relevance in others, so isolated visual inspection may miss heterogeneous decision strategies. Relevance maps therefore need to be examined at both the sample level and the dataset level. At the sample level, the aim is to determine whether a specific prediction is supported by meaningful regions; at the dataset level, the aim is to determine whether similar relevance patterns recur across many predictions. This local-to-global analysis is important because shortcut learning is not always visible in a single heatmap but may become evident when repeated relevance patterns are compared across samples. Relevance-based explainability should thus not be limited to producing heatmaps; it should provide a structured way to compare, summarise, and interpret decision evidence across a model's predictions.

Based on this need, the present study proposes Relevance-Based Model Decision Explainability (ReMoDEx) as a structured framework for assessing the decision behaviour of trained image classifiers. ReMoDEx is not designed to improve classification accuracy; rather it examines whether a model's predictions are supported by meaningful regions or by shortcut-sensitive visual patterns. The framework takes a trained classifier and its prediction-specific relevance maps as its main inputs and converts these maps into structured decision-evidence representations.

1. First, the trained classifier performs inference and the target class for explanation is selected from the output.
2. Second, relevance maps are generated for the selected target class using local explainability methods.
3. Third, the generated heatmaps are standardised through resizing, normalisation, and vectorisation, so that each prediction is represented in a comparable decision-pattern format.
4. Fourth, similar relevance patterns are grouped to identify recurring decision behaviours across samples.
5. Fifth, each decision-pattern group is interpreted by examining whether the dominant relevance lies in task-relevant regions or in non-relevant regions such as borders, corners, background areas, or artifact-sensitive zones.

This moves the analysis from individual prediction explanation to model-level decision assessment, providing a direct way to detect candidate shortcut-learning and Clever Hans-type behaviour that may remain hidden under conventional metrics.

## 1.2 Chest X-ray classification as a case study

To demonstrate the practical use of ReMoDEx, this study adopts chest X-ray multiclass disease classification as a case-study application. Chest X-ray classification is a suitable problem because radiographic images may contain both clinically meaningful thoracic findings and non-clinical cues introduced by acquisition, preprocessing, scanner variation, image borders, or dataset composition. This is especially relevant in COVID-19 chest X-ray classification, where images collected from different sources may carry source-dependent patterns unrelated to pulmonary pathology. In the present experiment, the COVID-19 Radiography Database is used with four diagnostic classes: COVID-19,

Normal, Lung Opacity, and Viral Pneumonia. A VGG16-based convolutional classifier is trained on this task, and the trained model is then used as the subject for ReMoDEx-based decision assessment. The purpose is not only to evaluate classification performance but to examine whether the decisions are supported by anatomically meaningful thoracic regions or by candidate shortcut-sensitive regions. Here, thoracic and lung-field relevance is treated as more consistent with task-relevant decision evidence, whereas excessive relevance on image borders, corners, peripheral structures, or non-pulmonary regions is treated as a candidate indicator of shortcut-sensitive behaviour. Model relevance concentrated within the thoracic and lung-field regions represents a more anatomically plausible decision pattern, whereas relevance shifted toward image borders, corners, and non-pulmonary artifacts represents a candidate shortcut-sensitive or Clever Hans-type pattern. The chest X-ray experiment thus provides a controlled setting to show how ReMoDEx can move beyond accuracy-based evaluation and identify heterogeneous decision strategies. It serves as the experimental demonstration of the proposed framework rather than as a standalone classification-performance study.

### 1.3 Objectives of the study

- To propose ReMoDEx as a relevance-based local-to-global model-decision explainability framework for assessing trained image classifiers.
- To combine each local explainability method with a single global relevance-pattern analysis module independently, so that every method produces its own set of decision-strategy results that can then be compared across methods.
- To examine whether a model's decision evidence is concentrated on task-relevant regions or shifted toward non-relevant regions.
- To provide a scalable assessment procedure for large-scale image datasets, where inspecting explanation heatmaps one sample at a time and human verifictaion is impractical.
- To demonstrate the practical use of ReMoDEx through a VGG16-based chest X-ray multiclass classification case study.
- To show that conventional performance metrics may be insufficient for identifying heterogeneous or shortcut-sensitive decision strategies.
- To validate different artifact-induced conditions in order to assess changes in model decision strategy.

### 1.4 Key contributions

- This study proposes ReMoDEx, a structured relevance-based assessment framework that moves beyond single-heatmap inspection and enables systematic interpretation of model decision behaviour.
- ReMoDEx combines each of four complementary local explainability methods with the same global relevance-pattern analysis module independently.
- The framework offers a scalable route to model explainability on large-scale datasets: the global module compresses an entire collection of relevance maps into a small number of decision-strategy clusters.
- The framework provides a procedure for identifying whether decisions are supported by task-relevant regions or influenced by shortcut-sensitive regions.
- The framework is demonstrated through a VGG16-based chest X-ray multiclass case study, showing that a classifier with acceptable conventional performance may still exhibit heterogeneous decision strategies.
- The study incorporates artifact-induced (masked-image) validation to assess whether changes in central or peripheral regions alter prediction confidence, relevance distribution, and decision behaviour.

## 2. Related work

The studies summarised in **Table 1** collectively show that shortcut learning, Clever Hans-type behaviour, and artifact-sensitive decision-making are important concerns in medical image analysis (Geirhos et al., 2020; Lapuschkin et al., 2019; Roberts et al., 2021). DeGrave et al. (2021) first provided strong explainability-based evidence in COVID-19 chest X-ray classification, showing that radiographic AI models may rely on source-specific and non-pathological cues rather than true disease-related evidence, which supports the concern that high apparent performance may still reflect confounding factors instead of clinically meaningful decision logic. Teixeira et al. (2021) examined this issue in COVID-19 and pneumonia classification by combining lung segmentation with explainability-based evaluation, showing that segmentation and explanation analysis help inspect whether attention lies inside clinically relevant regions, but that source-related bias may remain even after segmentation—restricting the image area alone may not fully eliminate shortcut-sensitive learning. Bassi and Attux (2022) extended this concern by using Layer-wise Relevance Propagation with external validation, showing that performance can drop under external testing and that relevance analysis is useful for assessing whether decisions are supported by symptomatic lung regions or by non-lung/background information. Jiménez-Sánchez et al. (2023) provided artifact-specific evidence of shortcut learning in pneumothorax classification, demonstrating that models may exploit non-pathological cues such as chest drains that become strongly associated with the target label.

Brown et al. (2023) broadened the discussion by proposing shortcut testing for fair medical AI across radiology and dermatology, showing that shortcut learning can affect not only reliability but also fairness when decisions are influenced by sensitive-attribute-related or confounded cues. Arias-Londoño and Godino-Llorente (2024) specifically analysed the Clever Hans effect in COVID-19 detection using interpretability and uncertainty-based analysis, reporting that classifiers can be influenced by spurious correlations while lung-region guidance reduces the effect—directly aligning with the present study, where relevance is examined to determine whether decisions are supported by thoracic regions or shortcut-sensitive areas. Bassi et al. (2024) further showed that background and non-target-region bias can drive shortcut learning across chest X-ray and other classification tasks, and that relevance-based optimisation can reduce background reliance and improve generalisation. Ong Ly et al. (2024) demonstrated that hidden acquisition bias can affect multimodal medical AI including X-ray and CT models, confirming that shortcut learning is not limited to a single dataset or disease category. Lin et al. (2024) extended shortcut-learning analysis beyond classification to medical image segmentation, showing that caliper annotations, padding, and cropping artifacts can influence outputs. Overall, these studies establish that medical imaging models may rely on source-specific cues, background regions, acquisition artifacts, annotation marks, or preprocessing patterns instead of intended clinical evidence, a problem that broader analyses of clinical AI reliability, interpretability, and trustworthiness have also emphasised (Topol, 2019; Kelly et al., 2019; Rudin, 2019; Oakden-Rayner et al., 2020; Ghassemi et al., 2021; Reyes et al., 2020). However, most existing work focuses on selected explanations, dataset bias, external validation, or specific artifact testing. There remains a need for a structured relevance-based, local-to-global decision-explainability framework that converts prediction-specific relevance maps into interpretable decision-pattern representations. The present study addresses this gap by proposing ReMoDEx and demonstrating it through a VGG16-based chest X-ray multiclass case study.

| Study | Modality / task | Explainability / assessment focus | Shortcut / bias examined | Main finding relevant to this study |
|---|---|---|---|---|
| DeGrave et al. (2021) | Chest X-ray; COVID-19 detection | Explainable AI analysis of COVID-19 CXR classifiers | Source-specific and non-pathological cues | COVID-19 radiographic models could rely on confounding factors rather than pathology, producing apparently accurate but unreliable models. |
| Teixeira et al. (2021) | Chest X-ray; COVID-19 / pneumonia | Lung segmentation with explanation-based evaluation | Dataset-source bias and non-lung evidence | Lung segmentation and explanation analysis are important, but segmented images may still retain source-related bias. |
| Bassi and Attux (2022) | Chest X-ray; COVID-19 classification | Relevance propagation with external validation | Non-lung/background contribution and dataset bias | External-validation performance dropped versus internal testing; relevance analysis helped assess whether decisions were supported by symptomatic lung regions. |

| Study | Modality / task | Explainability / assessment focus | Shortcut / bias examined | Main finding relevant to this study |
|---|---|---|---|---|
| Jiménez-Sánchez et al. (2023) | Chest X-ray; pneumothorax classification | Shortcut validation using annotated artifacts | Chest drains and artifact-associated labels | Models may exploit artifacts such as drains, supporting the need to assess image-based shortcuts beyond accuracy. |
| Brown et al. (2023) | Radiology and dermatology clinical AI | Shortcut testing for fairness assessment | Sensitive-attribute-related shortcut learning | Direct shortcut testing showed that shortcut learning can contribute to unfair or unreliable medical AI behaviour. |
| Arias-Londoño & Godino-Llorente (2024) | Chest X-ray; COVID-19 detection | Interpretability, uncertainty, and Clever Hans analysis | Spurious correlations in COVID-19 models | COVID-19 classifiers were affected by Clever Hans-type behaviour, and lung-region guidance reduced the effect. |
| Bassi et al. (2024) | Chest X-ray and other classification tasks | Relevance-optimised training to reduce background bias | Background and non-target-region bias | Background features can drive shortcut learning; relevance-based optimisation reduced background reliance and improved external generalisation. |
| Ong Ly et al. (2024) | Multimodal medical AI incl. X-ray and CT | Generalisation analysis under hidden acquisition bias | Data-acquisition bias | Hidden acquisition biases can overestimate performance by about 20%, showing that shortcut learning directly affects generalisation. |
| Lin et al. (2024) | Medical image segmentation | Shortcut identification and mitigation | Caliper annotations, padding, cropping artifacts | Extended shortcut-learning analysis beyond classification, showing annotation and preprocessing artifacts can affect segmentation outputs. |

**Table 1.** *Recent explainability- and shortcut-learning-focused studies demonstrating artifact-sensitive or Clever Hans-type model behaviour in medical imaging.*

# 3. Mathematical preliminaries

This section formalises the two components that ReMoDEx operates on: the trained VGG16 multiclass classifier that produces predictions, and the local explainability operators together with the global relevance-pattern analysis that convert those predictions into structured decision evidence. The notation introduced here is used throughout the remainder of the paper.

## 3.1 VGG16-based multiclass classifier

### 3.1.1 Dataset and input representation

Let the chest X-ray dataset be defined as $\mathcal{D} = \{(x_i, y_i)\}_{i=1}^{N}$ $(i = 1, \ldots, N)$, where $x_i \in \mathbb{R}^{H\times W\times C}$ denotes the $i$-th input image and $y_i \in \{0,1\}^K$ is the corresponding one-hot label. For the four-class task $K = 4$, corresponding to COVID-19, Normal, Lung Opacity, and Viral Pneumonia. Each image is min–max normalised before being passed to the network:

$$\tilde{x}_i = \frac{x_i - x_{\min}}{x_{\max} - x_{\min}}$$

### 3.1.2 Feature extraction

The VGG16 backbone is a composition of five convolutional blocks, $F_i = f_\theta(x_i) = (B_5 \circ B_4 \circ B_3 \circ B_2 \circ B_1)(x_i)$, where $\theta$ are the convolutional parameters. For the $l$-th convolutional layer the pre-activation feature map is

$$Z_{p,q,k}^{(l)} = b_k^{(l)} + \sum_{c,u,v} W_{u,v,c,k}^{(l)}\ A_{p+u,\,q+v,\,c}^{(l-1)}$$

with ReLU activation $A_{p,q,k}^{(l)} = \sigma\big(Z_{p,q,k}^{(l)}\big)$, $\sigma(z) = \max(0, z)$, and spatial down-sampling by max-pooling $P_{p,q,k}^{(l)} = \max_{(u,v)\in\Omega} A_{p+u,\,q+v,\,k}^{(l)}$. The final convolutional tensor is $F_i \in \mathbb{R}^{H\prime\times W\prime\times C\prime}$.

### 3.1.3 Classification head and prediction

The feature tensor is vectorized by $h_i = \phi(F_i)$; with global average pooling the $c$-th component is $h_{i,c} = \frac{1}{H'W'} \sum_{p,q} F_{i,p,q,c}$. The head maps the representation to a hidden embedding $g_i = \sigma(W_1 h_i + b_1)$ and then to class logits $s_i = W_o g_i + b_o$. The posterior of class $k$ follows the softmax

$$p_{ik} = P(y_i = k \mid x_i) = \frac{\exp(s_{ik})}{\sum_j \exp(s_{ij})}$$

and the predicted label is $\hat{y}_i = \text{argmax}_k p_{ik}$. Equivalently, the complete classifier is $\hat{y}_i = \text{argmax}_k \text{softmax}\Big(W_o\, \sigma \big(W_1\, \phi(f_\theta(x_i))\big)\Big)_k$.

### 3.1.4 Objective and optimisation

For one-hot labels the multiclass cross-entropy loss is $\mathcal{L}_{CE} = -\frac{1}{N}\sum_{i,k} y_{ik} \log p_{ik}$. Under class imbalance a weighted variant $\mathcal{L}_{WCE} = -\frac{1}{N}\sum_{i,k} \alpha_k\; y_{ik} \log p_{ik}$ is used, with inverse-frequency weights $\alpha_k = \frac{N}{KN_k}$. The regularised objective is $J(\Theta) = \mathcal{L}_{WCE} + \frac{\lambda}{2} \parallel \Theta \parallel_2^2$, minimised by $\Theta^* = \text{argmin}_\Theta J(\Theta)$ using Adam updates over the trainable parameters $\Theta = \{\theta, W_1, b_1, W_o, b_o\}$.

## 3.2 Local explainability operators

After training, local explainability is applied to the same model output to obtain class-specific relevance maps. For image $x_i$, the target class is $c_i = \hat{y}_i = \text{argmax}_k p_{ik}$, and the target-class score is $q_i(x) = f_\Theta^{(c_i)}(x)$. For a local method $m$ the relevance map is $M_i^m(c_i) = \Psi_m(x_i, f_\Theta, q_i)$, $M_i^m \in \mathbb{R}^{H\times W}$. Four complementary operators $\Psi_m$ are used.

### 3.2.1 Grad-CAM++

Let $A_i^r$ be the $r$-th activation map of the last convolutional layer and $G_{uv}^{r,c} = \frac{\partial q_i}{\partial A_{i,uv}^r}$ the class-specific gradient. The pixel-wise weighting coefficient is

$$\alpha_{uv}^{(r,c)} = \frac{(G_{uv}^{r,c})^2}{2(G_{uv}^{r,c})^2 + \sum_{a,b} A_{i,ab}^r\,(G_{uv}^{r,c})^3 + \varepsilon}$$

with importance weight $w_r^c = \sum_{u,v} \alpha_{uv}^{(r,c)}\; \rho(G_{uv}^{r,c})$, $\rho(z) = \max(0, z)$. The localisation map $L_i^{GC++} = \rho(\sum_r w_r^c A_i^r)$ is min–max normalised and resized to the input size to give the final map $M_i^{GC++}(c_i) = \uparrow_{H,W} L_i^{GC++}$.

### 3.2.2 Integrated Gradients

Integrated Gradients integrates the target-class gradient along a straight path from a baseline $x_i^0$ to $x_i$. For feature index $d$,

$$IG_{i,d}^c = \big(x_{i,d} - x_{i,d}^0\big) \int_0^1 \frac{\partial q_i\big(x_i(\alpha)\big)}{\partial x_{i,d}}\, d\alpha$$

approximated over $T$ steps as $IG_{i,d}^c \approx \frac{x_{i,d} - x_{i,d}^0}{T} \sum_t \frac{\partial q_i(x_i(t/T))}{\partial x_{i,d}}$. Spatial relevance follows by positive-sum channel aggregation, $L_i^{IG} = \sum_{ch} I\, G_{i,u,v,ch}^c$, followed by min–max normalisation to $M_i^{IG}$.

### 3.2.3 Occlusion Sensitivity

Occlusion sensitivity measures the target-class score drop when a local patch is masked. With mask $O_{uv}$, the occluded image is $x_i^{(u,v)} = x_i \odot (1 - O_{uv}) + x_i^0 \odot O_{uv}$, and the relevance is $L_i^{Occ}(u, v) = \max\Big(0,\; q_i(x_i) - q_i\big(x_i^{(u,v)}\big)\Big)$. After min–max normalisation and resizing this yields $M_i^{Occ}$.

### 3.2.4 Layer-wise Relevance Propagation

Layer-wise Relevance Propagation decomposes the target-class score backward through the network, initialising $R_k^{(L)} = q_i$ for $k = c_i$ and 0 otherwise. With contribution $z_{jk}^{(l)} = a_j^{(l)} w_{jk}^{(l,l+1)}$, the $\varepsilon$-stabilised rule propagates

$$R_j^{(l)} = \sum_k \frac{z_{jk}^{(l)}}{\sum_{j'} z_{j'k}^{(l)} + \varepsilon \cdot \text{sign}\left(\sum_{j'} z_{j'k}^{(l)}\right)} R_k^{(l+1)}$$

This approximately conserves relevance across layers, $\sum_j R_j^{(l)} \approx \sum_k R_k^{(l+1)}$. Positive-sum channel aggregation and min–max normalisation give the input-level map $M_i^{LRP}$.

### 3.3 Global relevance-pattern analysis

The global module of ReMoDEx operates on the standardised relevance maps produced by each local operator. For a given method the maps are resized, normalised, and vectorised into a feature matrix; a similarity graph is then constructed, its graph Laplacian is spectrally decomposed to estimate the number of recurring patterns, and the samples are partitioned into decision-pattern clusters using spectral clustering (Ng et al., 2002; von Luxburg, 2007). This global stage follows the local-to-global relevance-analysis principle introduced by Lapuschkin et al. (2019).

#### 3.3.1 Pairwise distance and similarity graph

For method $m$, let $z_i^m$ and $z_j^m$ be two vectorised heatmaps. The squared Euclidean distance is $d_{ij}^m = \| z_i^m - z_j^m \|_2^2 = \sum_r \left(z_{ir}^m - z_{jr}^m\right)^2$. Distances are mapped to similarities with a Gaussian kernel,

$$S_{ij}^m = \exp\left(-\frac{d_{ij}^m}{2\sigma^2}\right) = \exp\left(-\frac{\| z_i^m - z_j^m \|_2^2}{2\sigma^2}\right)$$

giving the similarity matrix $S^m \in \mathbb{R}^{N \times N}$; in practice a $k$-nearest-neighbour graph is used so that $S^m$ is sparse.

#### 3.3.2 Degree matrix and graph Laplacian

The degree matrix is $D_{ii}^m = \sum_j S_{ij}^m$, $D_{ij}^m = 0$ for $i \neq j$. The unnormalised Laplacian is $L^m = D^m - S^m$, and the normalised Laplacian is

$$L_{norm}^m = (D^m)^{-1/2} L^m (D^m)^{-1/2} = I - (D^m)^{-1/2} S^m (D^m)^{-1/2}$$

#### 3.3.3 Eigengap-based cluster selection

The normalised Laplacian is decomposed as $L_{norm}^m v_k^m = \lambda_k^m v_k^m$, with eigenvalues ordered $0 = \lambda_1^m \leq \lambda_2^m \leq \cdots \leq \lambda_N^m$. The eigengap is $\Delta_k^m = \lambda_{k+1}^m - \lambda_k^m$, and the number of clusters is selected as $K^{*m} = \text{argmax}_k \Delta_k^m$. A larger eigengap indicates stronger separation between recurring relevance-pattern groups.

#### 3.3.4 Spectral embedding and clustering

The first $K^{*m}$ eigenvectors form the embedding $U^m = [v_1^m, \ldots, v_{K^*}^m] \in \mathbb{R}^{N \times K^*}$, whose $i$-th row $u_i^m$ is the spectral representation of sample $i$. The embedded points are partitioned by $k$-means, $C^m = k\text{-means}(U^m, K^{*m})$, minimising $\sum_k \sum_{u \in C_k} \| u_i^m - \mu_k^m \|_2^2$ with centroids $\mu_k^m = \frac{1}{|C_k^m|} \sum u_i^m$. For visualisation the feature matrix is projected to two dimensions using principal component analysis.

## 4. The proposed ReMoDEx framework

### 4.1 Four independent local-to-global explanation channels

A central design decision in ReMoDEx is that each of four local explainability methods is connected to the global relevance-pattern analysis module independently, so that every local method forms its own self-contained local-to-global explanation channel rather than being averaged or fused with the others. The motivation is that no single local method is a complete or artifact-free description of model behaviour. Each method probes a different mechanism of the network and carries its own characteristic blind spots, so a decision pattern that appears under one method alone could reflect the explainer rather than the classifier. The four methods were therefore chosen because they are complementary in what they measure:

1. **Grad-CAM++** is a gradient-weighted, class-discriminative method computed from the last convolutional layer. It captures where high-level, semantically abstract features fire, producing coarse localisation of the regions most responsible for the class score. Its resolution is limited by the convolutional feature-map size and it can smooth relevance toward object boundaries.
2. Integrated Gradients is a path-integral input-attribution method that satisfies the sensitivity and implementation-invariance axioms. It gives fine-grained, pixel-level attributions but depends on the choice of baseline image, which can shift the apparent importance of low-intensity regions.
3. **Occlusion Sensitivity** is a perturbation-based, model-agnostic method that directly tests causality: it measures how much the target-class score drops when a region is physically removed. It is intuitive and faithful to the model's input–output behaviour, but its output depends on patch size and stride and is comparatively expensive.
4. **Layer-wise Relevance Propagation** is a conservation-based decomposition that redistributes the output score back onto the input while approximately preserving total relevance. It provides a theoretically grounded, whole-network attribution, but its result depends on the chosen propagation rule.

Because these four families are grounded in gradients, path integrals, perturbation, and relevance conservation respectively, they rarely fail in the same way on the same input. In ReMoDEx each local method is therefore passed through the identical global relevance-pattern module on its own, producing four independent local-to-global instantiations of the framework: Grad-CAM++ with the global module, Integrated Gradients with the global module, Occlusion Sensitivity with the global module, and Layer-wise Relevance Propagation with the global module. Each pairing is individually meaningful—on its own it converts a set of prediction-specific relevance maps into an interpretable set of decision-strategy clusters (for example, a central thoracic strategy and a border/corner strategy) for that one explainer—so a useful decision-level result is obtained even from a single channel. The scientific strength of ReMoDEx then comes from reading the four independent channels together. If a border- or corner-dominant decision structure were an artifact of one explainer, it would not be expected to recur across all four; conversely, when the same central-versus-peripheral organisation of relevance emerges independently in every channel, it is far more plausibly a genuine property of the trained model. Because all four channels are standardised into the same heatmap representation before clustering, their outputs are directly comparable, and their convergence (or divergence) provides the robust, method-agnostic evidence that lets ReMoDEx separate reproducible model-level decision strategies from explainer-specific noise.

## 4.2 Framework overview

Figure 1 illustrates the overall workflow of ReMoDEx. The pipeline begins with the chest X-ray dataset, image preprocessing, stratified data splitting, training of the VGG16-based multiclass classifier, and evaluation of baseline performance. After training, the four local explainability methods—Grad-CAM++, Integrated Gradients, Occlusion Sensitivity, and Layer-wise Relevance Propagation—are applied independently to generate class-specific relevance maps. Crucially, the relevance maps of each method are then carried through the global relevance-pattern analysis module separately, so that the global stage is instantiated once per local method and each local–global channel produces its own decision-strategy clusters. Within each channel the heatmaps are resized, vectorised, and normalised, then passed to the global module, where spectral clustering with eigengap-based selection identifies clusters of recurring explanation patterns. Each cluster is interpreted as a model decision strategy by comparing central thoracic relevance with edge/corner relevance, and masked-image validation is used to test whether predictions are sensitive to pulmonary or non-pulmonary regions. Because the global module can be applied efficiently to a whole large-scale set of relevance maps, each channel yields a more complete picture of classifier behaviour than a single heatmap, and the agreement of the four independent channels reveals unexpected or Clever Hans-type decision-making in a way that no single explainer could establish alone.

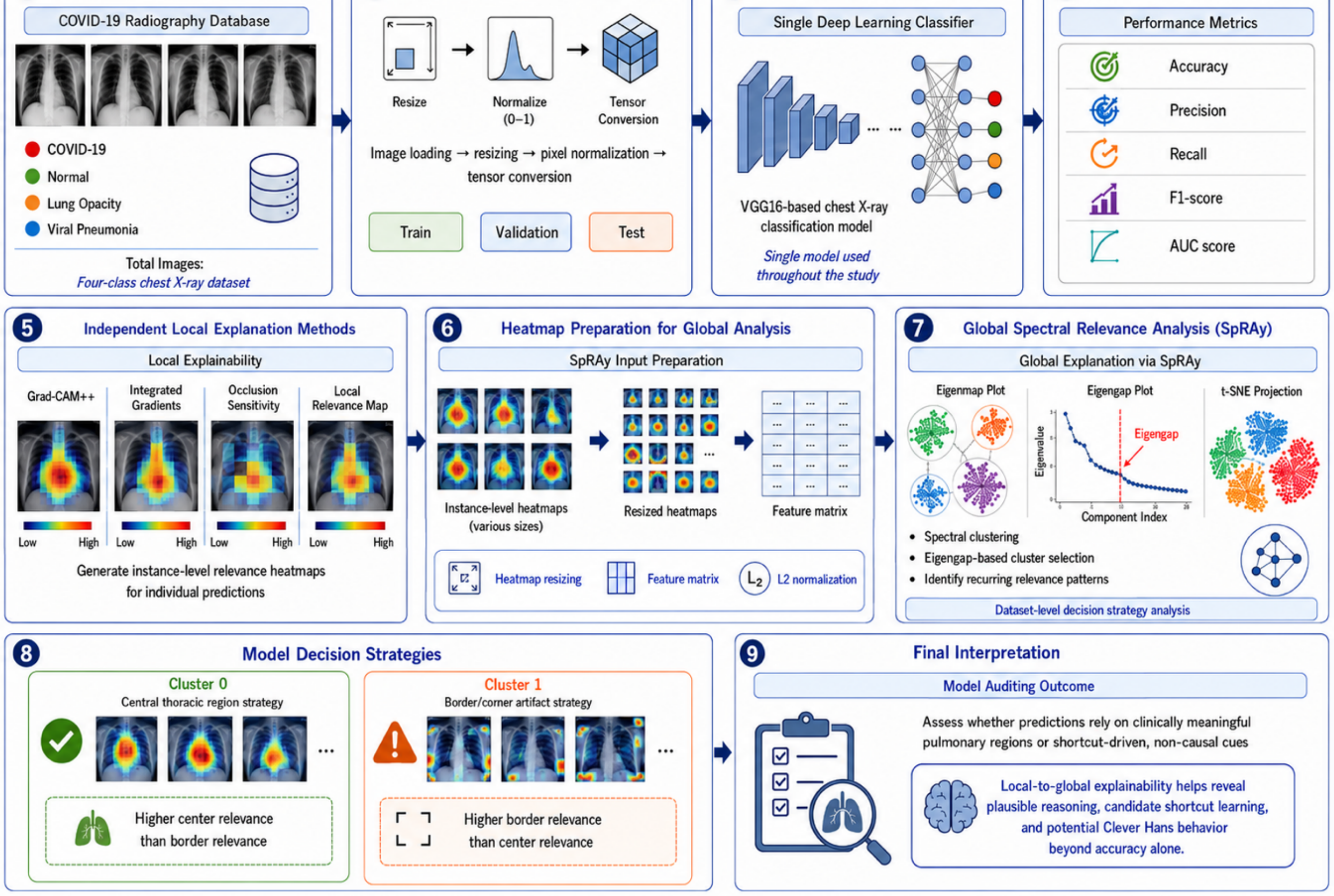

***Fig. 1.*** *Methodological workflow of ReMoDEx for detecting Clever Hans effects and shortcut learning in COVID-19 chest X-ray classification: from the CXR dataset and VGG16 training, through four independent local explainers, to global relevance-pattern clustering, decision-strategy interpretation, and masked-image validation.*

## 4.3 The ReMoDEx algorithm

The proposed algorithm implements a local-to-global explainability workflow whose theoretical basis model inference, relevance-map generation, similarity-graph construction, graph-Laplacian computation, eigengap analysis, and spectral clustering was formalised in Section 3. In operation the procedure is as follows. Step 1: each preprocessed chest X-ray image is passed through the trained VGG16 classifier to obtain class probabilities and the predicted target class. Step 2: the four local explainers independently generate class-specific relevance maps. Step 3: each heatmap is resized, vectorised, and organised into a method-specific feature matrix. Step 4: the heatmap vectors are mapped to a similarity graph based on pairwise distances. Step 5: spectral graph analysis with eigengap-based selection determines the number of recurring relevance-pattern clusters. Step 6: heatmaps are assigned to method-specific clusters by spectral clustering. Step 7: clustered heatmaps are visualised and quantified with spatial relevance measures. Step 8: each cluster is interpreted as a model decision strategy for example, central thoracic relevance or border/corner-dominant relevance followed by masked-image validation in the experimental phase. The algorithm thus bridges individual local explanations and global decision-pattern analysis, providing the procedural basis for identifying candidate shortcut-learning or Clever Hans-type behaviour in a trained classifier. Algorithm 1 summarises the steps and their mathematical operations.

| Steps | Operation, formulation, and expected output |
|---|---|
| **Input** | Trained classifier $f_\theta$; test images $\{x_i\}$ $(i = 1, \dots, N)$; four local methods $M = \{\mathrm{GC}^{++}, \mathrm{IG}, \mathrm{Occ}, \mathrm{LRP}\}$; number of classes $K = 4$. |
| **Output** | Method-specific global clusters $\{C^{\mathrm{GC}^{++}}, C^{\mathrm{IG}}, C^{\mathrm{Occ}}, C^{\mathrm{LRP}}\}$ and decision strategies $\{S^{\mathrm{GC}^{++}}, S^{\mathrm{IG}}, S^{\mathrm{Occ}}, S^{\mathrm{LRP}}\}$. |

| Steps | Operation, formulation, and expected output |
|---|---|
| **1. Model inference** | For each $x_i$ compute $s_i = f_\Theta(x_i)$. Output: class-logit vector $s_i$. |
| | $p_{ik} = \frac{\exp(s_{ik})}{\sum_j \exp(s_{ij})}$. Output: class probabilities $p_{ik}$. |
| | $c_i = \text{argmax}_k p_{ik}$; target score $q_i = s_{i,c_i}$. Output: target class and score. |
| **2. Local explanation** | For each $m \in M$ generate $M_i^m(c_i) = \Psi^m(x_i, f_\Theta, q_i) \in \mathbb{R}^{H\times W}$. Output: $M_i^{\text{GC}^{++}}, M_i^{\text{IG}}, M_i^{\text{Occ}}, M_i^{\text{LRP}}$. |
| **3. Heatmap prep** | $M_i^m \rightarrow \text{resized}(h \times w)$; $z_i^m = \text{vec}(M_i^m) \in \mathbb{R}^{hw}$. Output: vectorised heatmap $z_i^m$. |
| | $Z^m = [z_1^m; \ldots; z_N^m] \in \mathbb{R}^{N\times hw}$. Output: method-specific input matrix $Z^m$. |
| **4. Similarity graph** | $d_{ij}^m = \| z_i^m - z_j^m \|^2$. Output: distance matrix. |
| | $S_{ij}^m = \exp\left(-d_{ij}^m/(2\sigma^2)\right)$. Output: similarity matrix $S^m \in \mathbb{R}^{N\times N}$ (kNN-sparse). |
| **5. Spectral analysis** | $D_{ii}^m = \sum_j S_{ij}^m$; $L^m = D^m - S^m$; $L_{\text{norm}}^m = (D^m)^{-1/2} L^m (D^m)^{-1/2}$. |
| | Solve $L_{\text{norm}}^m v_k^m = \lambda_k^m v_k^m$; $\Delta_k^m = \lambda_{k+1}^m - \lambda_k^m$; $K^{*m} = \text{argmax}_k \Delta_k^m$. Output: cluster count $K^{*m}$. |
| **6. Clustering** | $U^m = [v_1^m, \ldots, v_{K^*}^m]$; $C^m = \text{k} - \text{means}(U^m, K^{*m})$. Output: cluster set $C^m$. |
| **7. Visualisation** | 2D projection of $Z^m$ (PCA / t-SNE). Output: cluster visualisation. |
| **8. Interpretation** | For each cluster inspect heatmaps and spatial relevance; assign $C_k^m \Rightarrow S_k^m$ (e.g. central thoracic vs. border/corner). Output: decision strategy $S^m$. |

***Table 2.*** *Algorithm 1( local-to-global relevance-based decision assessment in ReMoDEx).*

## 4.4 Computational complexity analysis

The cost of ReMoDEx separates cleanly into a local relevance-generation stage and a global relevance-pattern analysis stage. Let N be the number of assessed samples, d = h×w the flattened heatmap dimension, K* the selected number of clusters, and let F denote the cost of one forward–backward pass through the trained network. The four local methods are applied independently, so the constant factor of the local stage is four.

**Local stage:** Grad-CAM++ and LRP each require essentially one forward pass and one backward/relevance pass, so their per-sample cost is O(F). Integrated Gradients evaluates T points along the integration path, costing O(T·F) per sample. Occlusion Sensitivity slides a patch of size p with stride s, producing $P = ((H-p)/s + 1)^2$ masked forward passes per sample, costing $O(P\cdot F_{\text{fwd}})$. Summing over N samples, the local stage is O( N·F·(2 + T + P) ), dominated in practice by Integrated Gradients (the T factor) and Occlusion Sensitivity (the P factor), which are the two most expensive explainers.

**Global stage (per method):** Vectorisation of N heatmaps costs O(N·d). Building a full similarity matrix costs $O(N^2\cdot d)$; with a k-nearest-neighbour graph this reduces to a sparse structure with O(N·k) edges. Forming the graph Laplacian is O(N·k) for the sparse case. The eigendecomposition of the normalised Laplacian is $O(N^3)$ in the dense case, but only the first K* eigenvectors are needed, so an iterative solver on the sparse graph reduces this to roughly O(N·k·K* ) per iteration. The k-means step on the K*-dimensional embedding costs $O(N\cdot K^{*2}\cdot t_{iter})$, and the two-dimensional PCA projection costs about O(N·d). The dominant term is therefore the graph construction and eigen-solution: $O(N^2\cdot d + N^3)$ in the dense worst case, reduced to $O(N\cdot k\cdot d + N^2\cdot K^*)$ with the sparse kNN graph and a partial eigensolver.

**Overall:** Because the four methods share the same global pipeline, the total time complexity is $O(4\cdot[N\cdot F\cdot(2 + T + P) + N^2\cdot d + N^3])$, which simplifies to $O(N\cdot F\cdot(T + P) + N^2\cdot d + N^3)$. Space complexity is dominated by storing the heatmap feature matrix, O(N·d), together with the similarity matrix and Laplacian, $O(N^2)$ dense or O(N·k) sparse, and the eigenvectors O(N·K*), giving $O(N^2 + N\cdot d)$ overall. For the assessment sizes used here (N in the range of 100 relevance maps per method, d = 1024), the quadratic and cubic graph terms are small, so the runtime is in practice governed by the local stage—in particular the P forward passes of Occlusion Sensitivity and the T steps of Integrated Gradients. This is why ReMoDEx is intended as an offline assessment procedure applied to a representative sample of predictions rather than as an inference-time overhead on every deployment query. Crucially, the same design is what makes the framework suitable for large-scale datasets: because the global module compresses an arbitrarily large collection of relevance maps into a small, fixed number of decision-strategy clusters, the analyst reads a handful of cluster prototypes instead of inspecting tens of thousands of individual heatmaps. Manual, sample-by-sample verification does not scale to such datasets and is impractical and time-consuming, whereas ReMoDEx surfaces the model's dominant decision strategies automatically making it applicable precisely in the large-scale settings where human intervention on every prediction is infeasible.

# 5. Experimental setup

## 5.1 Dataset

The study uses the publicly available COVID-19 Radiography Database (Chowdhury et al., 2020; Rahman et al., 2021) for chest X-ray classification and explainability-based assessment. The dataset contains 21,165 chest X-ray images distributed across four diagnostic classes: 3,616 COVID-19, 10,192 Normal, 6,012 Lung Opacity, and 1,345 Viral Pneumonia. Class labels were taken from the original folder-level annotations provided by the dataset maintainers. Each image was treated as a two-dimensional radiograph and processed through a standardised pipeline of loading, resizing to the model input size, pixel-intensity normalisation, and conversion to a model-compatible tensor. The dataset was used to train and evaluate a single chest X-ray classifier, which was then analysed with the four independent local explanation methods and the global relevance-pattern analysis module.

## 5.2 Model implementation

All images were loaded as three-channel radiographs and resized to 224 × 224 × 3 to meet the VGG16 input requirement. Images were converted to floating-point tensors and preprocessed with the standard VGG16 preprocessing function; the same pipeline was applied consistently for training, validation, testing, and explainability analysis. The data were split into training, validation, and testing subsets in a 70:15:15 stratified ratio, giving 14,815 training, 3,175 validation, and 3,175 testing images. Reproducibility was improved by fixing the random seed to 42 for Python, NumPy, and TensorFlow, with class order COVID, Lung Opacity, Normal, Viral Pneumonia.

A transfer-learning classifier was built on VGG16 (Simonyan & Zisserman, 2015) initialised with ImageNet weights (Deng et al., 2009) (include_top = False); such transfer learning from large natural-image corpora is a standard and effective strategy for medical image classifiers (Shin et al., 2016; He et al., 2016). The convolutional base was frozen during training, and a custom head was added consisting of global average pooling, dropout at rate 0.30, and a final dense layer with four softmax outputs. The model was optimised with Adam (learning rate $1 \times 10^{-4}$) and sparse categorical cross-entropy loss, trained for up to 10 epochs with batch size 8. To address class imbalance, balanced class weights were computed from the training labels: 1.4634 for COVID, 0.8802 for Lung Opacity, 0.5192 for Normal, and 3.9318 for Viral Pneumonia. Training was monitored by validation loss with early stopping (patience 5) restoring the best weights, and a learning-rate-reduction callback that scaled the learning rate by 0.2 after three epochs without improvement, down to a minimum of $1 \times 10^{-7}$. The model was evaluated with accuracy, precision, recall, F1-score, and one-vs-rest multiclass AUC, reported in Section 6.

## 5.3 Explainability implementation

The same trained VGG16 classifier was used for both local and global explainability analysis. Four independent local methods—Grad-CAM++, Integrated Gradients, Occlusion Sensitivity, and Layer-wise Relevance Propagation—were implemented and run independently, producing class-specific relevance maps for each prediction. In every case the explanation target was the class predicted by the model. The resulting local heatmaps were then used as inputs to the global relevance-pattern analysis module.

For Grad-CAM++, the target layer was the last convolutional layer (block5_conv3). A balanced set of 100 test samples was used (25 per class). Heatmaps were generated at 224 × 224 and visualised with overlay transparency 0.35 using the magma colormap. For Integrated Gradients, a black baseline image was used over 100 test samples, with 50 integration steps and batch size 16 along the attribution path; attributions were aggregated by positive-sum channel aggregation and normalised before analysis. For Occlusion Sensitivity, patches of 32 × 32 pixels with stride 16 were applied over 100 test samples, and relevance was computed from the positive drop in the target-class score after occlusion; the resulting maps were normalised and resized. For Layer-wise Relevance Propagation, a corrected fast implementation with the z-plus/epsilon rule ($\varepsilon = 1 \times 10^{-6}$) and positive-sum channel aggregation was used over 100 selected test samples.

For the global relevance-pattern analysis, the module operated on the relevance maps rather than the original images, since the goal was to identify recurring explanation patterns and decision strategies. Each heatmap was resized to 32 × 32 and flattened into a 1,024-dimensional feature vector, and the features were L2-normalised before graph construction. A k-nearest-neighbour similarity graph with 15 neighbours was built, and the RBF sigma was estimated from the median positive neighbour distance. Spectral clustering was applied to the relevance-map feature matrix, with the number of clusters selected automatically by the eigengap method over a search range of 2–8 clusters. Two-dimensional visualisation used principal component analysis, and cluster separation was assessed with the silhouette, Davies–Bouldin, and Calinski–Harabasz scores. To interpret clusters, average heatmaps were scored for border, centre, upper, and lower relevance, peak-location measures, and normalised entropy.

For masked-image validation, Grad-CAM++ was applied under three input conditions: original image, border-masked image, and centre-masked image. In the border-masked condition the outer 12% of the image was occluded with a black mask; in the centre-masked condition a black mask covered the central 60%. To ensure a fair comparison across conditions, the masked-image Grad-CAM++ explanations were generated with the original predicted class fixed as the explanation target. This enabled the analysis to quantify changes in target-class support, prediction confidence, prediction change, and regional relevance distribution under masking.

# 6. Results

## 6.1 Baseline performance of the VGG16 classifier

The baseline VGG16 model achieved stable performance across the training, validation, and testing sets (Table 3), with training accuracy 86.63%, validation accuracy 85.92%, and testing accuracy 86.27%. The weighted precision, recall, F1-score, and AUC were closely matched across splits, indicating that the model generalised consistently without a major drop on unseen test data. This confirms that, judged purely by conventional metrics, the classifier would be considered a well-performing model—precisely the situation in which shortcut-sensitive behaviour is most likely to go unnoticed.

| Split | Accuracy | Precision | Recall | F1-score | AUC | Samples |
|---|---|---|---|---|---|---|
| Training | 0.8663 | 0.8670 | 0.8663 | 0.8661 | 0.9638 | 14,815 |
| Validation | 0.8592 | 0.8598 | 0.8592 | 0.8589 | 0.9609 | 3,175 |
| Testing | 0.8627 | 0.8635 | 0.8627 | 0.8625 | 0.9624 | 3,175 |

**Table 3.** *Classification performance of the baseline VGG16 model on the training, validation, and testing sets.*

## 6.2 Independent per-method relevance-pattern results

Each of the four local explainers was combined with the global relevance-pattern analysis module independently: a method was first analysed at the sample level, and its relevance maps were then passed through the global module on their own to produce that method's decision-strategy clusters. This yields four self-contained sets of results, reported per method below. Because the four channels are produced independently, the recurring central-versus-peripheral organisation of relevance that appears across them—rather than any single method's output is what the subsequent discussion draws on.

### 6.2.1 Grad-CAM++

At the sample level, Grad-CAM++ overlays for correctly classified representatives of all four classes—Viral Pneumonia (confidence 0.9980), Normal (0.6387), Lung Opacity (0.9931), and COVID-19 (0.8583)—concentrated relevance within the thoracic field, including the lung regions, central chest area, and disease-relevant pulmonary zones, representing an anatomically plausible decision pattern presented in Fig. 2. A second set of correctly classified samples—Viral Pneumonia (0.9576), Normal (0.8359), Lung Opacity (0.7300), and COVID-19 (0.6575) instead showed stronger activation in peripheral regions such as upper borders, side regions, and corners, with the central lung parenchyma less prominent than the periphery presented in Fig. 3. Thus, correct predictions were supported by different visual evidence: the model did not follow one uniform strategy but exhibited both anatomically plausible relevance and candidate shortcut-like relevance tied to image borders or acquisition-related structures.

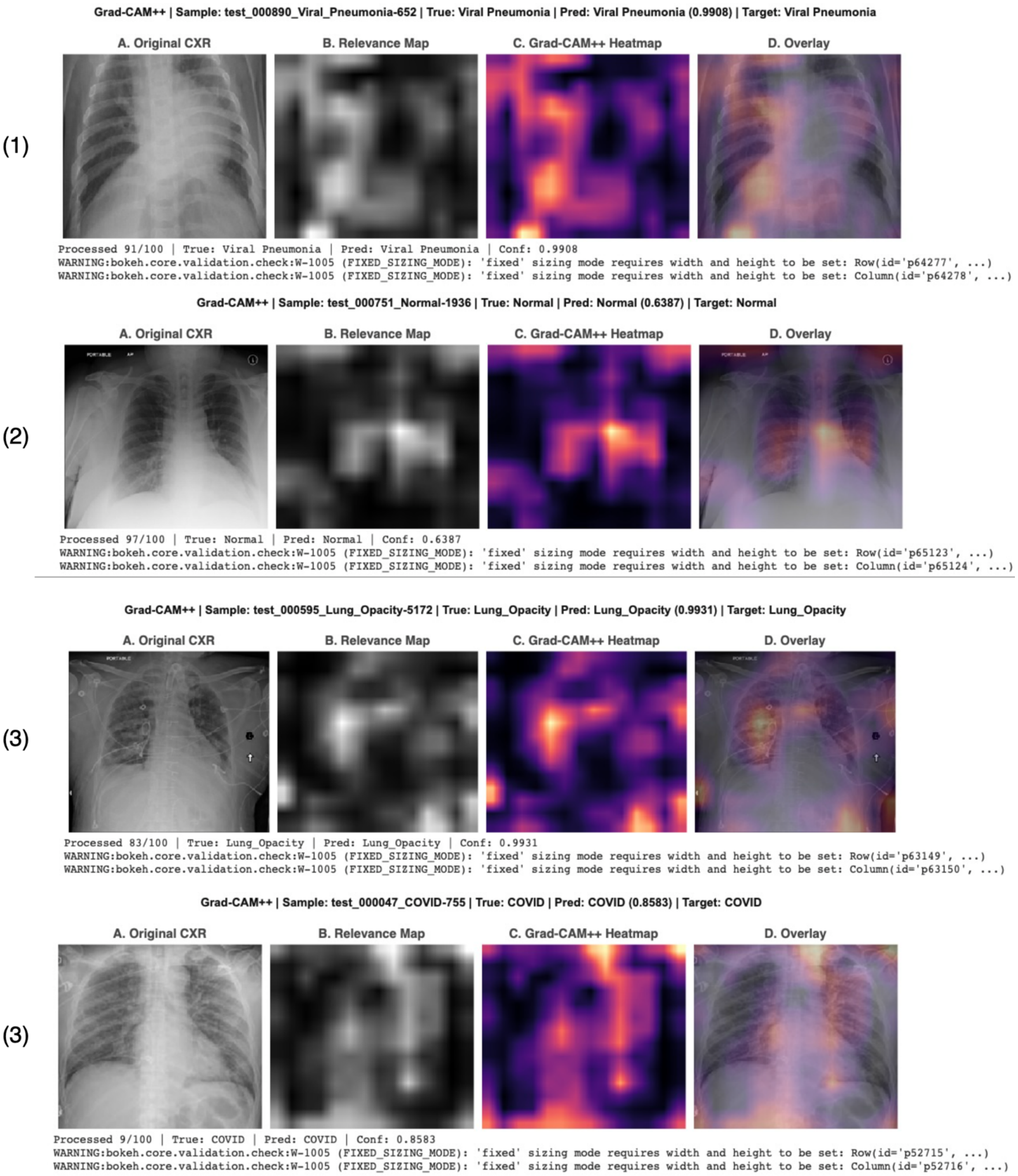

***Fig. 2. Grad*-CAM++-based assessment of anatomically plausible VGG16 decision behavior in chest X-ray classification, showing model relevance concentrated within thoracic and lung-field regions** *(Note: Each representative sample is presented using four image components: A. Original CXR, showing the input chest radiograph; B. Relevance Map, showing the grayscale spatial relevance distribution generated from the model response; C. Grad-CAM++ Heatmap, showing class-discriminative activation intensity using a color-coded heatmap; and D. Overlay, showing the Grad-CAM++ heatmap superimposed on the original CXR for visual localization of the highlighted regions.)*

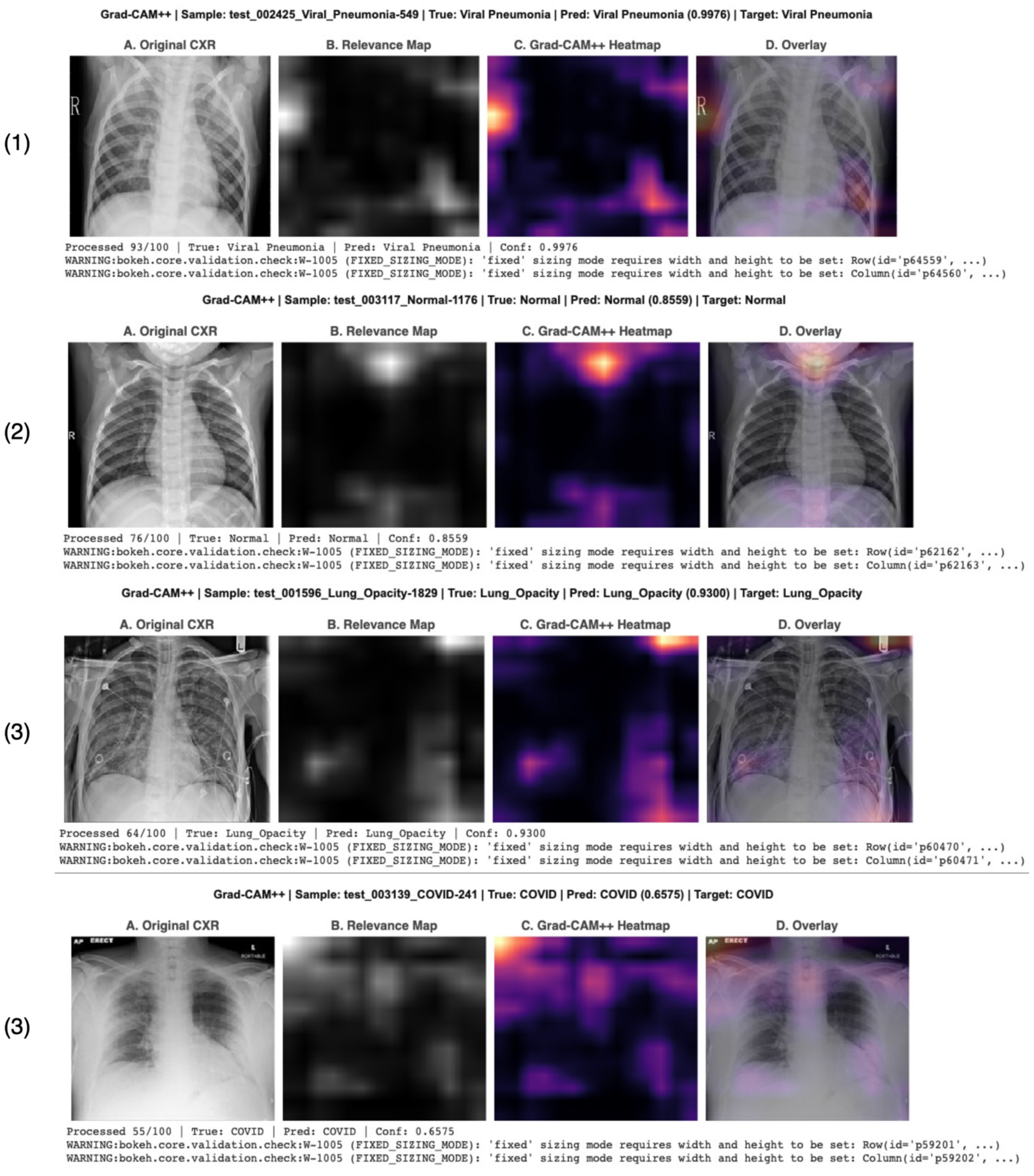

***Fig. 3**. **Grad-CAM++-based assessment of potential shortcut learning and Clever Hans-type VGG16 decision behavior in chest X-ray classification, showing model relevance concentrated around peripheral borders, corner regions, and non-pulmonary artifacts*** (Note: Each representative sample is presented using four image components: A. Original CXR, showing the input chest radiograph; B. Relevance Map, showing the grayscale spatial relevance distribution generated from the model response; C. Grad-CAM++ Heatmap, showing class-discriminative

activation intensity using a color-coded heatmap; and D. Overlay, showing the Grad-CAM++ heatmap superimposed on the original CXR for visual localization of the highlighted regions.)

At the dataset level, combining Grad-CAM++ with the global module independently separated the 100 Grad-CAM++ relevance maps into two decision-pattern clusters as shown in Fig. 4, with the eigengap criterion selecting two clusters (42 and 58 samples) as presented in Fig. 4 and Fig. 5 respectively. Global cluster-validity metrics showed modest separation: silhouette 0.0806, Davies–Bouldin 3.0024, and Calinski–Harabasz 10.2558 (Table 4). Cluster 0 was centrally organised, with centre relevance 0.5392 exceeding border relevance 0.2973, and was interpreted as a possible central thoracic-region strategy (mean confidence 0.8256, accuracy 0.8333). Cluster 1 showed the opposite pattern, with border relevance 0.5013 exceeding centre relevance 0.2628, interpreted as a possible border/corner artifact strategy (mean confidence 0.8647, accuracy 0.8448). The spatial measures in Table 5 and Table 6 confirm this: both clusters had balanced upper and lower relevance and high entropy, so the principal spatial difference was central versus peripheral rather than vertical. The class-distribution summary in Table 7 shows that Cluster 0 was mostly Normal (n = 17) and COVID (n = 14), while Cluster 1 contained more Viral Pneumonia (n = 21) and Lung Opacity (n = 18). The moderate label purities indicate that the clusters represent explanation-pattern groups rather than class-specific groups. The border-dominant cluster is a candidate shortcut pattern, but should be treated as suggestive rather than definitive evidence of Clever Hans behaviour without the masked-image validation reported in Section 6.3.

| Metric | Value |
|---|---|
| Number of samples | 100 |
| Number of clusters | 2 |
| Cluster selection method | Eigengap |
| Clustering backend | Spectral clustering |
| Feature normalisation | L2 |
| Heatmap size | 32 × 32 |
| Feature-matrix shape | 100 × 1024 |
| kNN neighbours | 15 |
| RBF sigma | 0.758787 |
| Silhouette score | 0.080647 |
| Davies–Bouldin score | 3.002377 |
| Calinski–Harabasz score | 10.255761 |
| PCA explained variance 1 | 0.129341 |
| PCA explained variance 2 | 0.080413 |

***Table 4.*** *Global relevance-pattern clustering metrics for Grad-CAM++.*

| Cl. | n | Suggested strategy | Dom. true | True purity | Dom. pred. | Pred. purity | Mean conf. | Acc. |
|---|---|---|---|---|---|---|---|---|
| 0 | 42 | Possible central thoracic-region strategy | Normal | 0.4048 | Normal | 0.5000 | 0.8256 | 0.8333 |
| 1 | 58 | Possible border/corner artifact strategy | Viral Pneumonia | 0.3621 | Viral Pneumonia | 0.3966 | 0.8647 | 0.8448 |

***Table 5.*** *Grad-CAM++ cluster decision-strategy summary.*

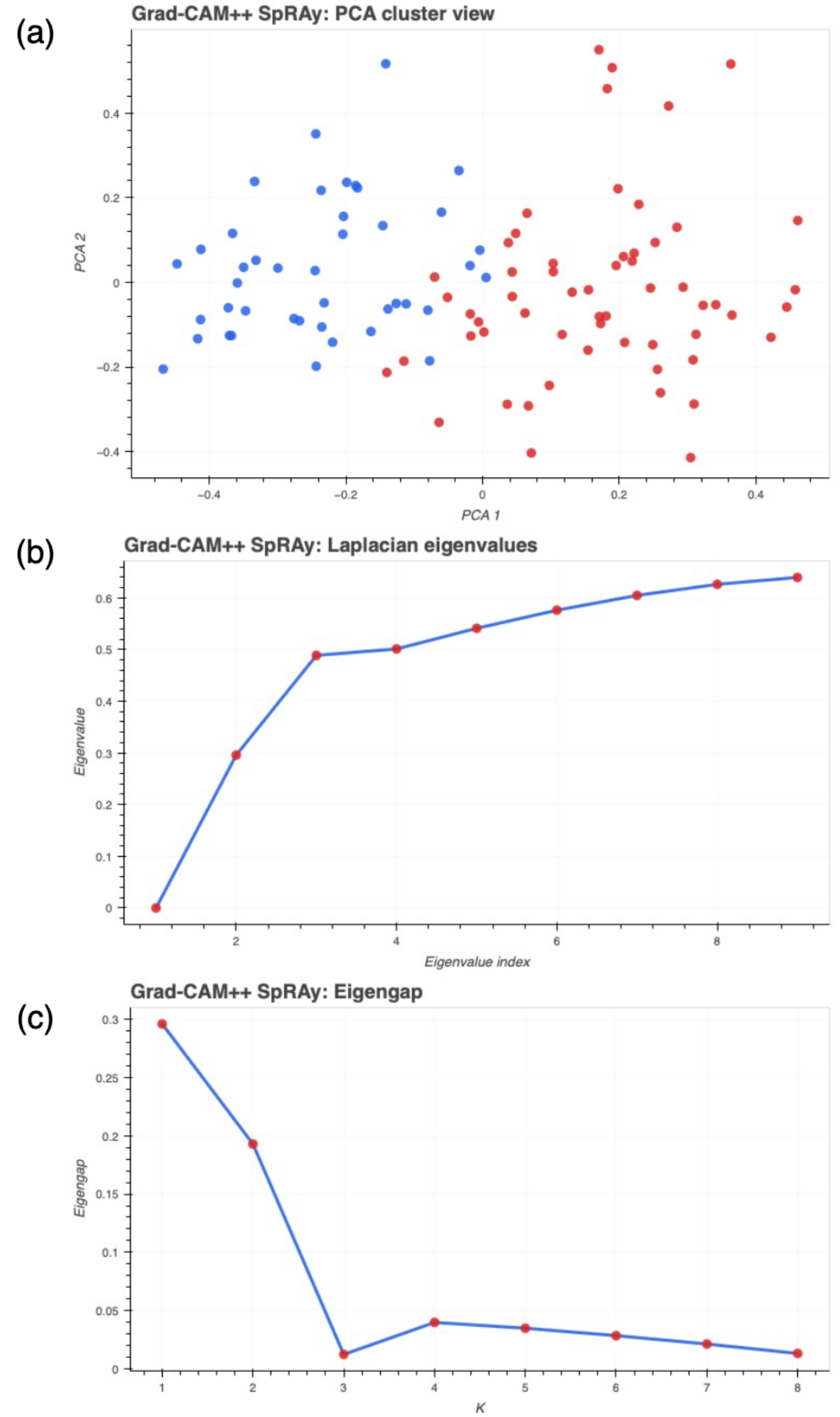

***Fig. 4.*** *Grad-CAM++-based clustering analysis of chest X-ray relevance maps for detecting dominant model decision strategies. Panel (a) visualizes the cluster separation in PCA space, panel (b) shows the Laplacian eigenvalue spectrum used for spectral clustering, and panel (c) shows the eigengap curve supporting the selection of two major explanation clusters.*

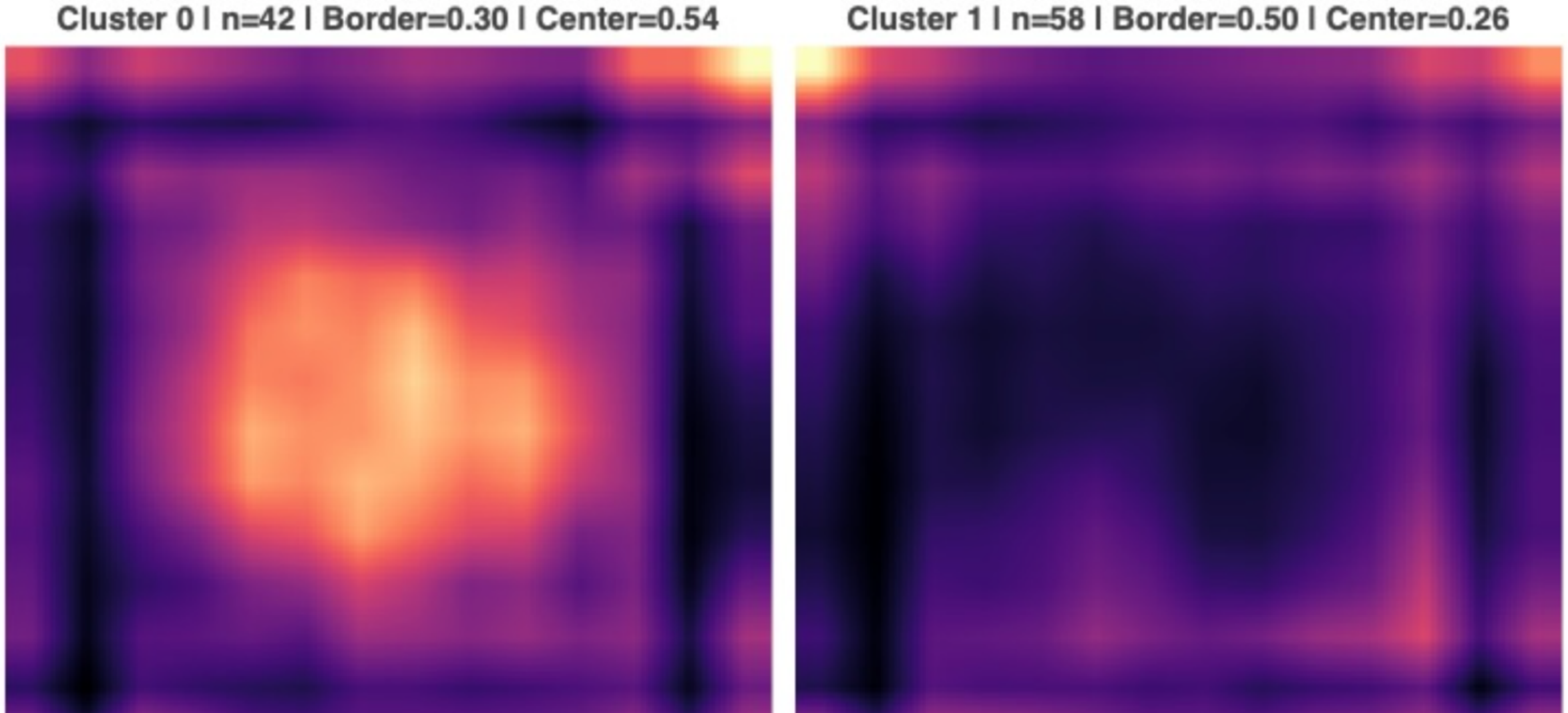

***Fig. 5. Grad-CAM++ ReMoDEx-derived decision strategies in chest X-ray classification.*** *The figure shows two dominant decision strategies identified by GLobal Relevance Analysis using Grad-CAM++ as the local explanation method. The left panel represents Cluster 0 / Strategy 1, where relevance is concentrated mainly in the central thoracic region, suggesting a more anatomically plausible decision pattern. The right panel represents Cluster 1 / Strategy 2, where relevance is concentrated more strongly along the image borders and corners, suggesting a candidate shortcut or Clever Hans-type decision pattern.*

| Cl. | n | Suggested strategy | Border | Centre | Upper | Lower | Peak Y | Peak X | Entropy |
|---|---|---|---|---|---|---|---|---|---|
| 0 | 42 | Central thoracic region | 0.2973 | 0.5392 | 0.5190 | 0.4810 | 0.0000 | 0.9686 | 0.9868 |
| 1 | 58 | Border/corner artifact | 0.5013 | 0.2628 | 0.5134 | 0.4866 | 0.0000 | 0.0000 | 0.9884 |

***Table 6.*** *Spatial relevance quantification of the Grad-CAM++ clusters.*

| Cl. | n | COVID (T) | L.Op. (T) | Norm. (T) | V.Pn. (T) | COVID (P) | L.Op. (P) | Norm. (P) | V.Pn. (P) |
|---|---|---|---|---|---|---|---|---|---|
| 0 | 42 | 14 | 7 | 17 | 4 | 14 | 3 | 21 | 4 |
| 1 | 58 | 11 | 18 | 8 | 21 | 15 | 16 | 4 | 23 |

***Table 7.*** *Class distribution across the Grad-CAM++ clusters (T = true label, P = predicted label).*

### 6.2.2 Integrated Gradients

Combined with the global module independently, Integrated Gradients split its 100 relevance maps into two clusters, with partial PCA separation Fig. 6(a) mainly along the first component: Cluster 0 spread toward negative PCA-1 and Cluster 1 toward positive PCA-1, with overlap near the centre indicating that the two groups were not fully separated. The cluster-level average maps showed that relevance was largely preserved in the centre for both clusters as shown Fig.6(b). Cluster 0 (47 samples) had centre relevance 0.50 and border relevance 0.28, a more centre-oriented pattern; Cluster 1 (53 samples) had centre relevance 0.45 and border relevance 0.32, with slightly more border involvement. Relative to Cluster 0, Cluster 1 reduced centre relevance by 0.05 and increased border relevance by 0.04. The main difference was therefore not a sharp central-versus-border division but a slight shift from a more central pattern in Cluster 0 to a slightly more peripheral pattern in Cluster 1. Overall, Integrated Gradients produced two related, largely central relevance-pattern groups with only partial separation.

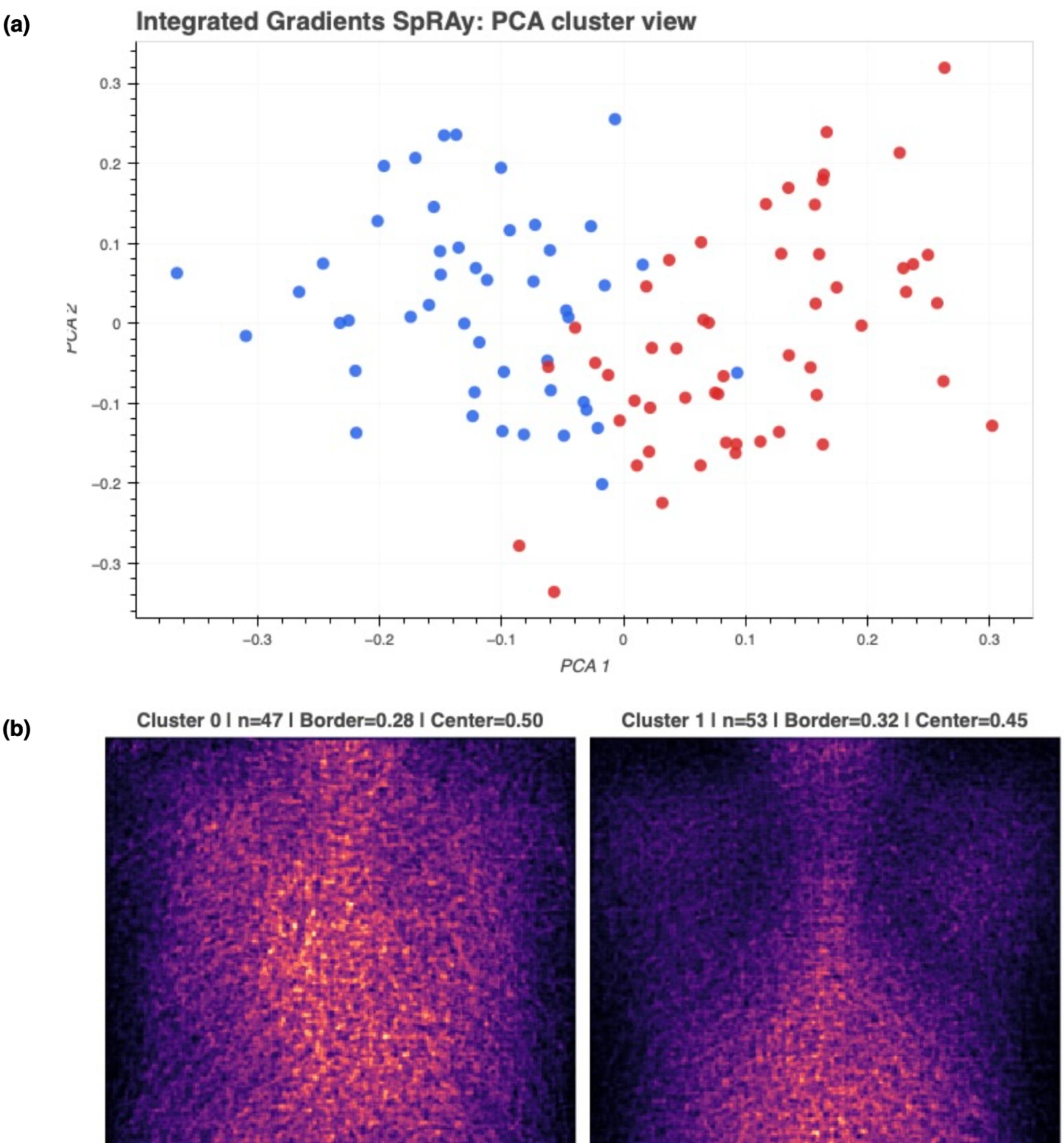

***Fig. 6*. *Integrated Gradients ReMoDEx-derived decision strategies in chest X-ray classification.*** *The figure shows two dominant decision strategies identified by Global Relevance Analysis using Integrated Gradients as the local explanation method. Panel* ***(a)*** *shows the PCA-based separation of explanation maps into two SpRAy clusters. Panel* ***(b)****shows the corresponding cluster prototypes, where the* ***left*** *panel represents* ***Cluster 0 / Strategy 1****, characterized by relatively stronger central relevance, and the* ***right*** *panel represents* ***Cluster 1 / Strategy 2****, characterized by comparatively stronger border-related relevance, suggesting a possible shortcut-sensitive decision pattern.*

### 6.2.3 Occlusion Sensitivity

Combined with the global module independently, Occlusion Sensitivity produced two clearly separated clusters among its 100 relevance maps. In PCA space Cluster 0 concentrated on the negative side and Cluster 1 on the positive side of the first component, with little overlap—the cleanest separation among the four methods as presented in Fig 7(a). In Fig 7(b), Cluster 0 (67 samples) showed a strongly central pattern, with centre relevance 0.61 and border relevance 0.17. Cluster 1 (33 samples) showed the opposite, with very high border relevance 0.81 and very low centre relevance 0.05, concentrating relevance on the image periphery, borders, and corners. The gap between clusters was large: Cluster 1 had 0.64 higher border relevance (0.81 − 0.17) and Cluster 0 had 0.56 higher centre relevance (0.61 − 0.05). Occlusion Sensitivity thus produced a well-separated central-thoracic pattern and a border-dominant pattern, giving the sharpest confirmation of the two-strategy structure seen across methods.

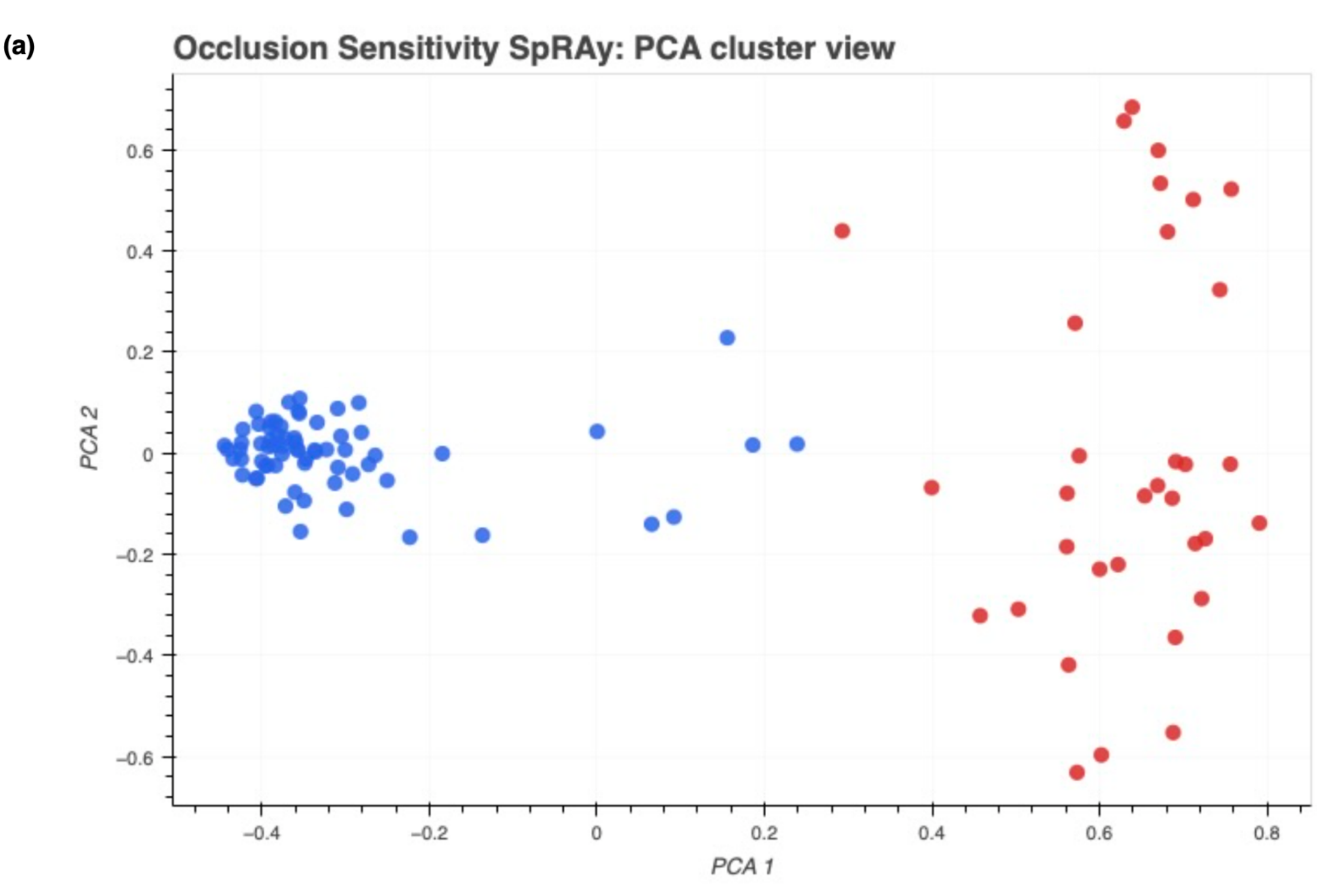

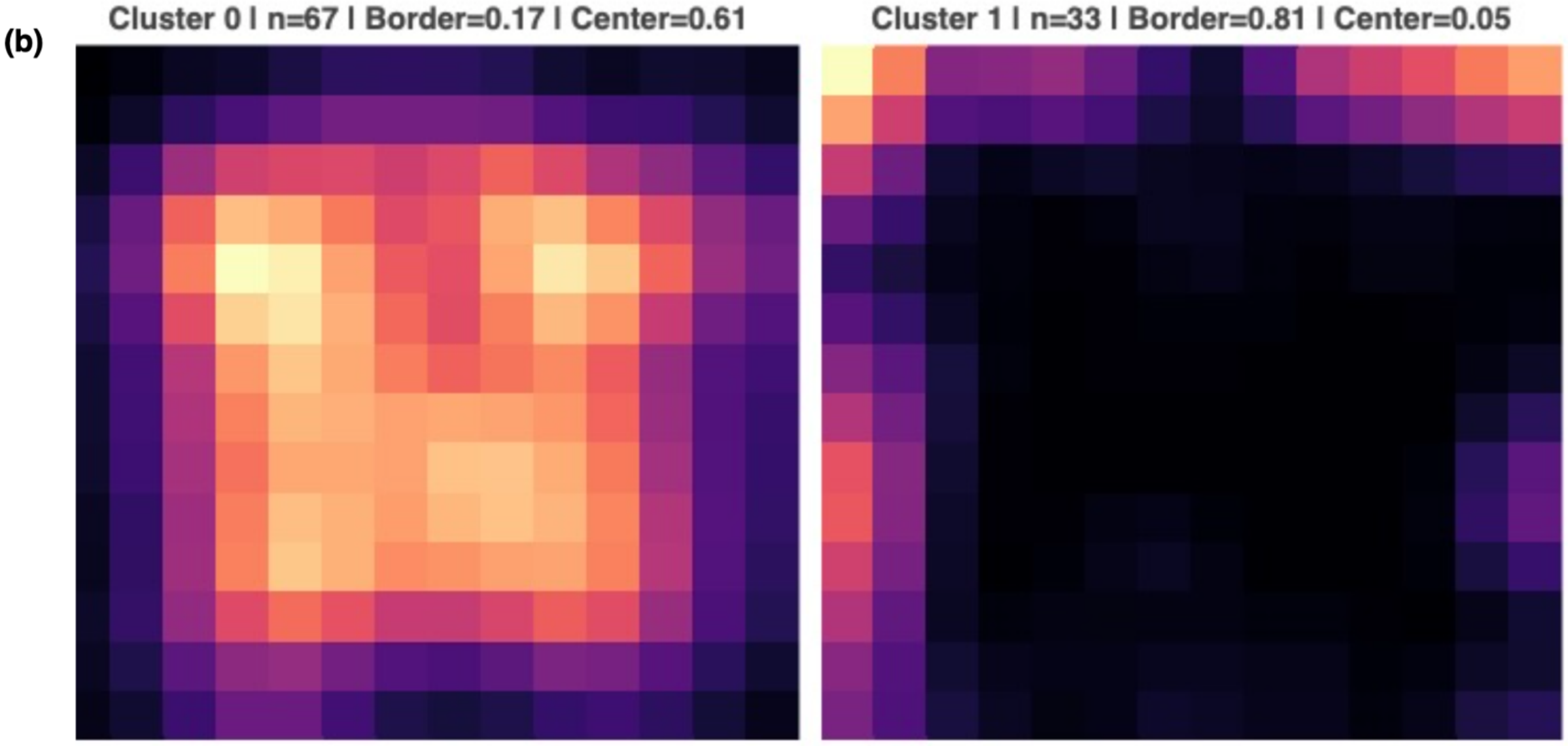

*Fig. 7.* ***Occlusion Sensitivity ReMoDEx-derived decision strategies in chest X-ray classification.*** *The figure shows two dominant decision strategies identified by Global Relevance Analysis using Occlusion Sensitivity as the local explanation method. Panel* ***(a)*** *shows the PCA-based separation of occlusion-derived explanation maps into two SpRAy clusters. Panel* ***(b)*** *shows the corresponding cluster prototypes, where the* ***left*** *panel represents* ***Cluster 0 / Strategy 1****, characterized by high central relevance and low border relevance, suggesting a more thoracic-*

*region-dependent decision pattern. The **right** panel represents **Cluster 1 / Strategy 2**, characterized by strong border-dominant relevance and very low central relevance, suggesting a candidate shortcut or Clever Hans-type decision pattern.*

### 6.2.4 Layer-wise Relevance Propagation

At the sample level, Layer-wise Relevance Propagation was examined for four representative test samples (two Lung Opacity, two Normal). In Fig. 6(1), a Lung Opacity sample was correctly classified with high confidence (0.9101), with relevance spread over the thoracic outline, the lower/basal lung areas, and peripheral structures rather than a single localised opacity.

In Fig. 6(2), a second Lung Opacity sample was misclassified as COVID with low confidence (0.4650), with relevance focused along lung boundaries, lower-thorax contours, and peripheral structures. Both Normal samples were correctly classified: one with moderate confidence (0.6888) and relevance along the lung margins and thoracic contours, and one with very high confidence (0.9968) showing an extensive bilateral pattern along rib-like and lung-boundary regions. These observations indicate that relevance was frequently distributed along anatomical borders and peripheral thoracic structures in both correctly and incorrectly classified samples.

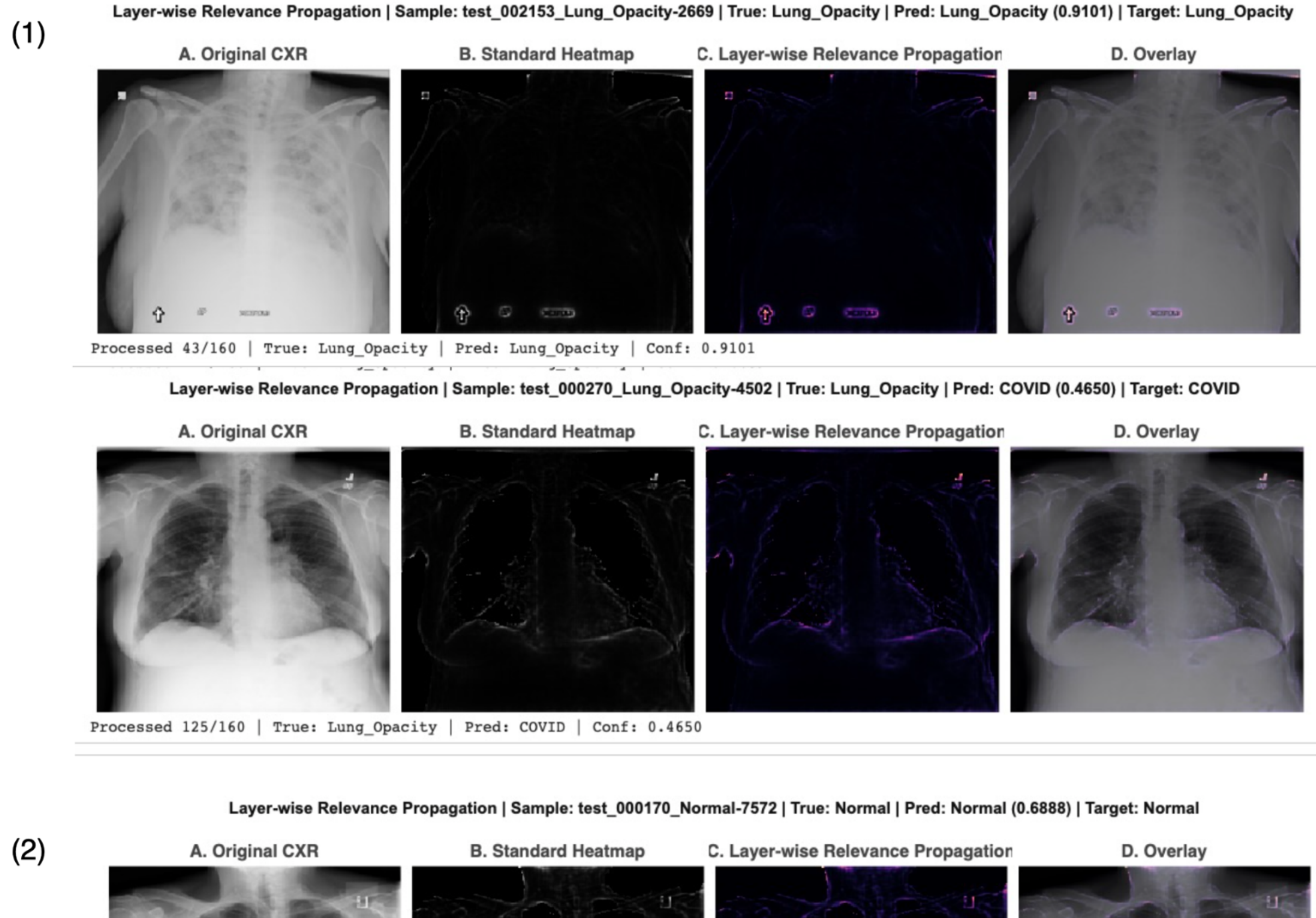

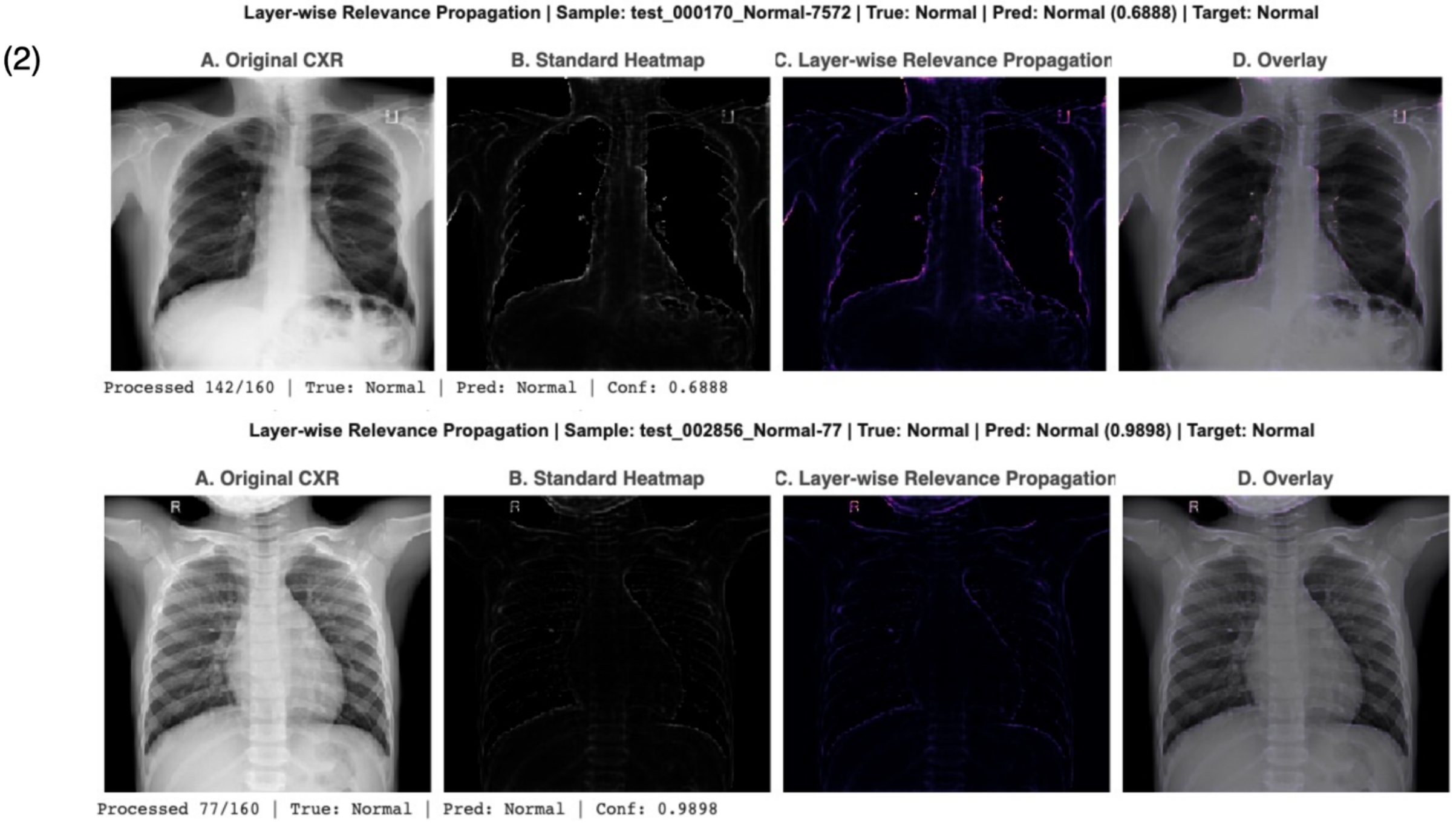

***Fig. 8. Local relevance-map visualization using Layer-wise Relevance Propagation for chest X-ray classification.****The figure presents representative local explanation outputs for individual chest X-ray predictions. Each row includes the* ***original CXR***, ***standard heatmap***, ***Layer-wise Relevance Propagation map***, *and* ***overlay visualization***. *Panel* ***(1)****shows Lung Opacity cases, including examples where the model either correctly predicts Lung Opacity or misclassifies the image as COVID. Panel* ***(2)*** *shows Normal cases where the model predictions are supported by relevance patterns distributed mainly along thoracic and lung-related structures.*

At the dataset level, combining Layer-wise Relevance Propagation with the global module independently partitioned the 100 LRP relevance maps into two clusters, with a strongly unbalanced split: Cluster 0 with 144 samples and Cluster 1 with 16 samples. In PCA space Cluster 0 occupied the main region while Cluster 1 lay toward positive PCA-1 with some overlap Fig 9(1). The Laplacian eigenvalue spectrum rose sharply for the first indices and then more gradually, supporting the two-cluster structure as presented in Fig 9(2). The cluster-mean maps showed in Fig 9(3), that both clusters carried peripheral and thoracic-outline relevance rather than purely central localisation: Cluster 0 was diffuse, with border relevance 0.41 slightly exceeding centre relevance 0.36, while Cluster 1 was sparser and more border-oriented, with border relevance 0.47 and centre relevance 0.32. LRP thus yielded one large cluster with diffuse thoracic-and-peripheral relevance and one smaller, more border-oriented cluster, consistent with the sample-level finding that the model often assigned relevance to lung borders, thoracic contours, and peripheral regions rather than restricting it to central lung parenchyma.

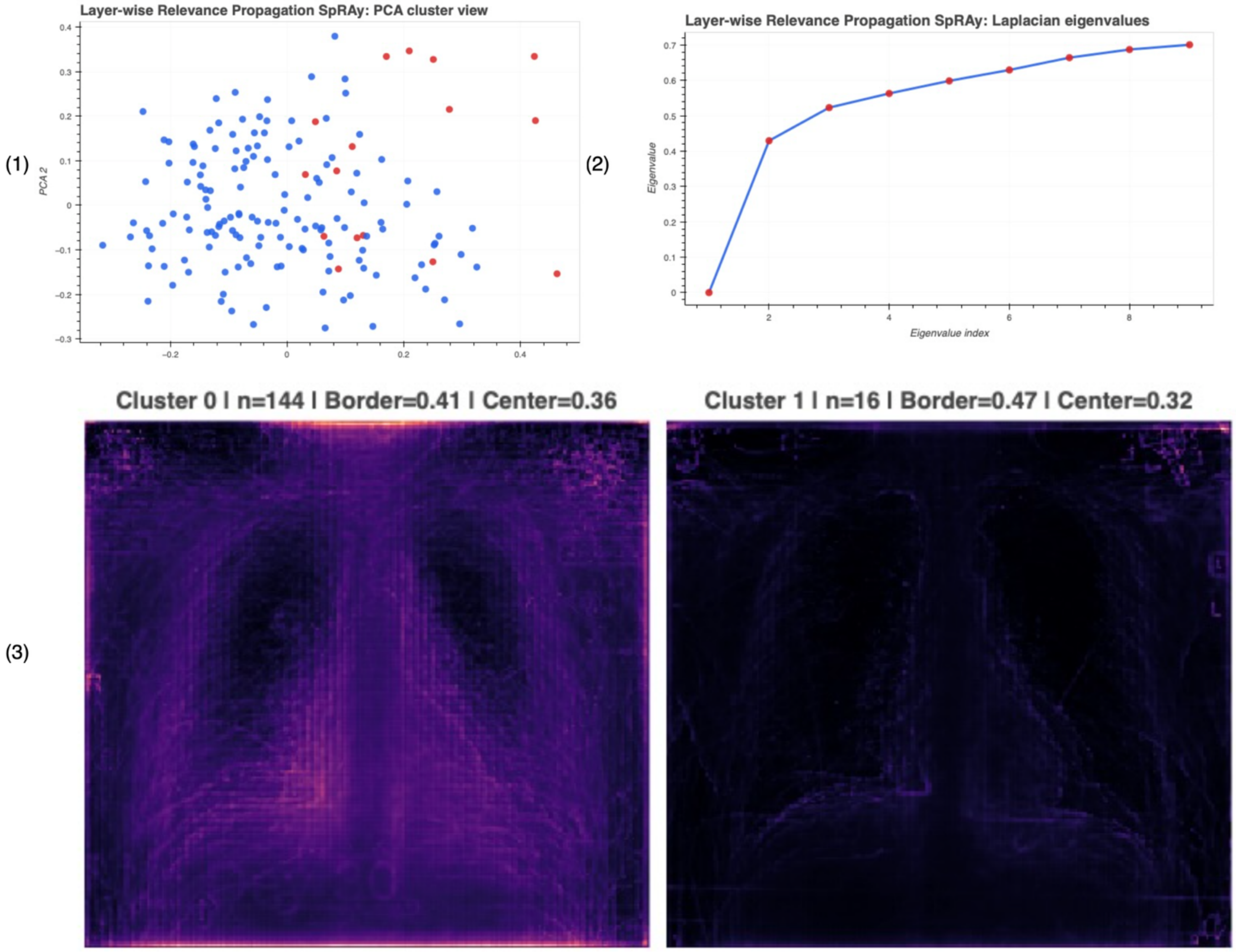

**Fig. 9.** ***LRP-based ReMoDEx analysis of chest X-ray relevance maps for identifying dominant model decision strategies.*** *The figure shows the Global Relevance Analysis results obtained using Layer-wise Relevance Propagation as the local explanation method. Panel* ***(1)*** *visualizes the PCA-based separation of LRP relevance maps into two SpRAy clusters, panel* ***(2)*** *shows the Laplacian eigenvalue spectrum used for spectral clustering, and panel* ***(3)*** *presents the two cluster prototypes. In panel* ***(3)****, the* ***left*** *image represents* ***Cluster 0 / Strategy 1****, showing broader thoracic and border-associated relevance, while the* ***right*** *image represents* ***Cluster 1 / Strategy 2****, showing a comparatively sparse relevance pattern with stronger peripheral/border influence, indicating two distinct LRP-derived decision strategies at the dataset level.*

## 6.3 Masked-image validation

The final and most direct test of decision behaviour was masked-image validation, in which Grad-CAM++ was recomputed under three conditions: original, border-masked (outer 12% removed) and centre-masked (central 60% removed)—with the original predicted class fixed as the explanation target. This validation is placed last because it converts the correlational cluster patterns of Section 6.2 into a causal test: if a region is genuinely used by the model, occluding it should degrade support for the original class.

### 6.3.1 Representative single-sample validation

For a representative Viral Pneumonia test image in Fig. 10, the original unmasked model correctly predicted Viral Pneumonia with confidence ≈0.91 and a distributed relevance pattern that contributed roughly equally from border and centre (border ≈0.33, centre ≈0.33) with balanced upper and lower ratios (≈0.50 each). Under border masking the prediction remained Viral Pneumonia but confidence fell to ≈0.67, with the fixed target score dropping to ≈0.67; relevance was largely retained in the visible thoracic region, particularly the lower-central lung area, so the remaining central information was sufficient to preserve the original decision at reduced confidence. Under centre masking the prediction changed from Viral Pneumonia to COVID (confidence ≈0.71), the fixed target score fell further to ≈0.36, and relevance shifted to the unmasked periphery (border ≈0.68, centre ≈0.16). This single sample shows a clear spatial sensitivity: the central thorax was needed to maintain the original decision, while the periphery alone was sufficient to push an alternative prediction once the centre was removed.

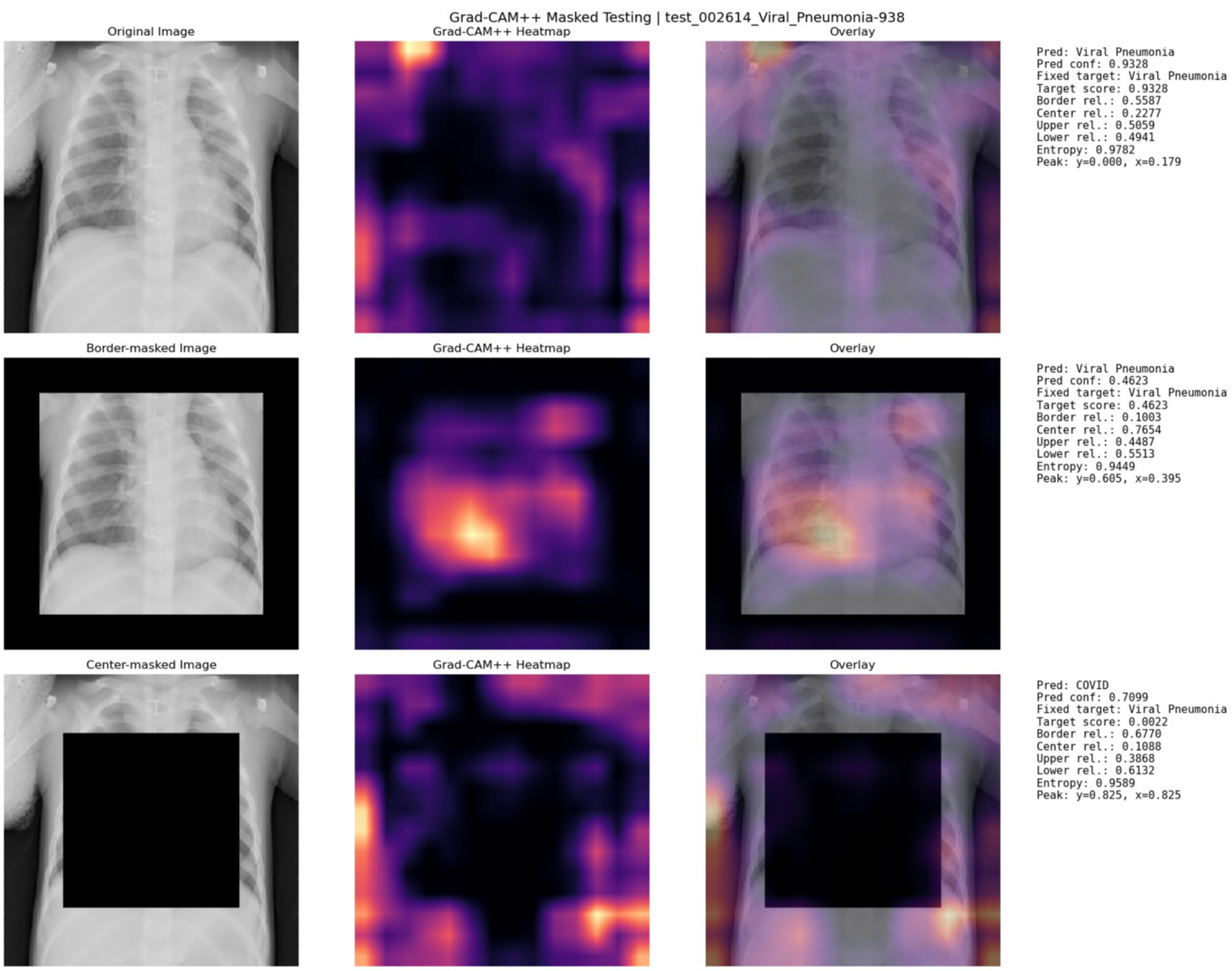

***Fig. 10. Sample-wise masked validation of Grad-CAM++ relevance patterns under original, border-masked, and center-masked chest X-ray conditions.*** *The figure presents a representative Viral Pneumonia sample evaluated under three input conditions: original image, border-masked image, and center-masked image. For each condition, the corresponding Grad-CAM++ heatmap and overlay visualization are shown to assess how regional masking alters the model's attention, confidence, and predicted class. The comparison demonstrates that masking central thoracic or peripheral border regions changes the relevance distribution and prediction behaviour, supporting sample-level evidence of region-dependent decision sensitivity.*

### *6.3.2 Cluster-level validation across conditions*

The condition-wise PCA projections in Fig. 11(a–c) provide the visual basis for comparing how Grad-CAM++ relevance-map clusters changed under original, border-masked, and centre-masked conditions. In the original condition, Fig. 11(a)shows two partially separated relevance-pattern groups, indicating the coexistence of central and peripheral decision strategies. After border masking, Fig. 11(b) shows a four-cluster structure with greater dispersion and class mixing, suggesting that removal of peripheral information altered the organisation of relevance patterns. In contrast, after centre masking, Fig. 11(c) again shows a two-cluster structure, but with a clearer shift toward peripheral/border-dependent relevance patterns. These visual trends support the subsequent cluster-level interpretation.

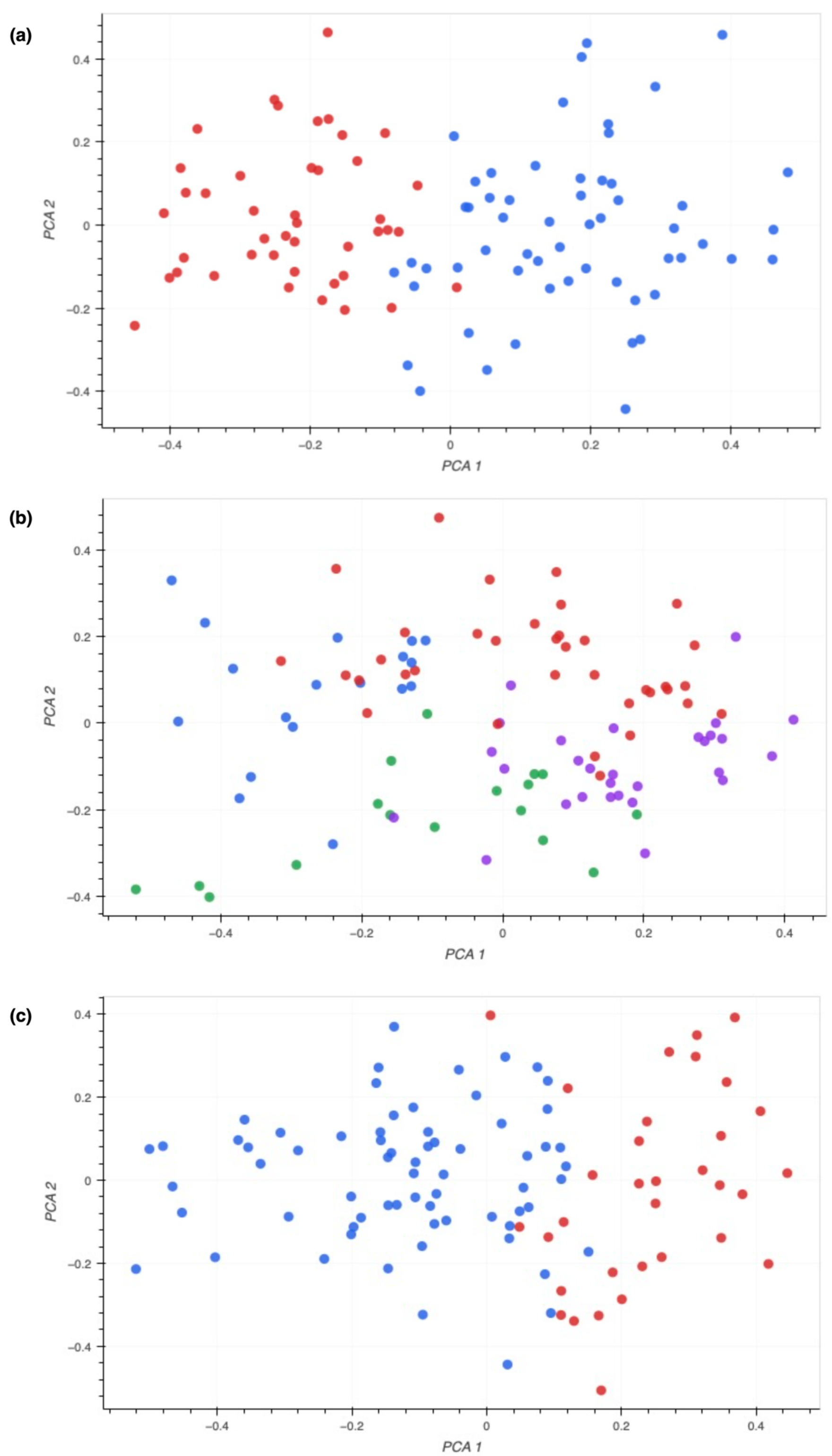

***Fig. 11. PCA-based visualization of relevance-map clustering under different masking conditions.***
***(a)*** *Original condition, showing clustering patterns derived from unmasked relevance maps, with two identified clusters: Cluster 0 = 57 samples and Cluster 1 = 43 samples.* ***(b)*** *Border-masked condition, showing redistribution of relevance-map patterns after masking peripheral/border regions, with four identified clusters: Cluster 0 = 18 samples, Cluster 1 = 36 samples, Cluster 2 = 17 samples, and Cluster 3 = 29 samples.* ***(c)*** *Center-masked condition, showing clustering patterns after masking the central thoracic region, with two identified clusters: Cluster 0 = 67 samples and Cluster 1 = 33 samples. Each point represents one sample projected onto the first two principal components, and colors indicate the identified clusters.*

At the cluster level in Fig. 12, the global module was applied to the masked-condition relevance maps. At the condition level, Table 8 shows that the original condition retained a high fixed target score of 0.8482, with no confidence drop and no prediction change, while the overall border/centre relevance ratio was 0.4167/0.3706. The clustering-quality measures in Table 9 further indicate modest cluster separation for the original condition, with silhouette 0.0837, Davies–Bouldin 2.9045, Calinski–Harabasz 10.9049, and mean entropy 0.9694.

| Condition | Clusters | Dominant relevance strategy | Fixed target | Conf. drop | Change rate | Border / Centre |
|---|---|---|---|---|---|---|
| Original | 2 | Possible border/corner artifact strategy | 0.8482 | 0.0000 | 0.00 | 0.4167 / 0.3706 |
| Border-masked (outer 12%) | 4 | Possible central thoracic-region strategy | 0.5073 | 0.3410 | 0.50 | 0.1086 / 0.7600 |
| Centre-masked (central 60%) | 2 | Possible border/corner artifact strategy | 0.3880 | 0.4602 | 0.55 | 0.5520 / 0.2199 |

***Table 8.*** *Prediction confidence, dominant relevance strategy, and border/centre relevance ratios under original and masked conditions (spectral-clustering backend; mean original prediction confidence ≈ 0.8483 in all conditions).*

| Condition | Clusters | RBF sigma | Silhouette | Davies–Bouldin | Calinski–Harabasz | Mean entropy |
|---|---|---|---|---|---|---|
| Original | 2 | 0.701414 | 0.0837 | 2.9045 | 10.9049 | 0.9694 |
| Border-masked (outer 12%) | 4 | 0.520427 | 0.0769 | 2.3687 | 9.3396 | 0.9349 |
| Centre-masked (central 60%) | 2 | 0.653011 | 0.0920 | 2.8618 | 10.3082 | 0.9563 |

***Table 9.*** *Spectral-clustering parameters, PCA-based cluster-validity metrics, and mean entropy under original and masked conditions.*

In the original condition it found two clusters: Cluster 0 (57 samples), dominated by Viral Pneumonia cases and predictions, was border-oriented (border 0.5013, centre 0.2614); Cluster 1 (43 samples), dominated by Normal, was central (centre 0.5151, border 0.3015). The unmasked model therefore already contained both a border/corner-oriented and a central thoracic-region strategy.

| Cl. | n | Suggested strategy | Dom. true | Dom. pred. | Fixed tgt. | Conf. drop | Change | Border | Centre |
|---|---|---|---|---|---|---|---|---|---|
| ***Condition = Original images*** | | | | | | | | | |
| 0 | 57 | Possible border/corner artifact strategy | Viral Pneumonia | Viral Pneumonia | 0.8694 | 0.0000 | 0.0000 | 0.5013 | 0.2614 |
| 1 | 43 | Possible central thoracic-region strategy | Normal | Normal | 0.8202 | 0.0000 | 0.0000 | 0.3015 | 0.5151 |
| ***Condition = Border-masked images (outer 12% removed)*** | | | | | | | | | |
| 0 | 18 | Possible central thoracic-region strategy | Lung Opacity | COVID | 0.4334 | 0.3572 | 0.5556 | 0.1216 | 0.6931 |
| 1 | 36 | Possible central thoracic-region strategy | Viral Pneumonia | COVID | 0.4620 | 0.4014 | 0.5278 | 0.1101 | 0.7636 |
| 2 | 17 | Possible central thoracic-region strategy | COVID | COVID | 0.6828 | 0.2313 | 0.3529 | 0.1082 | 0.7503 |
| 3 | 29 | Possible central thoracic-region strategy | Normal | COVID | 0.5063 | 0.3202 | 0.5172 | 0.0958 | 0.8031 |
| ***Condition = Centre-masked images (central 60% removed)*** | | | | | | | | | |

| Cl. | n | Suggested strategy | Dom. true | Dom. pred. | Fixed tgt. | Conf. drop | Change | Border | Centre |
|---|---|---|---|---|---|---|---|---|---|
| 0 | 67 | Possible border/corner artifact strategy | Normal | COVID | 0.4203 | 0.4057 | 0.5224 | 0.5307 | 0.2383 |
| 1 | 33 | Possible border/corner artifact strategy | Viral Pneumonia | COVID | 0.3225 | 0.5709 | 0.6061 | 0.5972 | 0.1738 |

***Table 10.*** *Cluster-wise relevance patterns and prediction behaviour under original and masked conditions. The eigengap criterion selected 2 clusters for the original condition, 4 for the border-masked condition, and 2 for the centre-masked condition. Fixed target score is the model score for the original target class under each condition; confidence drop is its reduction relative to the original prediction.*

When the border was masked, relevance shifted consistently toward the central thoracic region: four clusters were found, all with low border relevance (0.0958–0.1216) and high centre relevance (0.6931–0.8031).

This condition-level shift is also reflected in Table 8, where the border/centre relevance ratio changed to 0.1086/0.7600, confirming strong movement of relevance toward the central thoracic region. However, the fixed target score decreased to 0.5073, with a confidence drop of 0.3410 and an overall prediction-change rate of 0.50. The corresponding clustering metrics in Table 9 show four clusters with silhouette 0.0769, Davies–Bouldin 2.3687, Calinski–Harabasz 9.3396, and mean entropy 0.9349.

Despite this central shift, the dominant predicted label became COVID across all four clusters even though their dominant true labels were Lung Opacity, Viral Pneumonia, COVID, and Normal, and prediction-change rates rose to 0.3529–0.5556, removing the border concentrated relevance centrally but also destabilised the prediction.

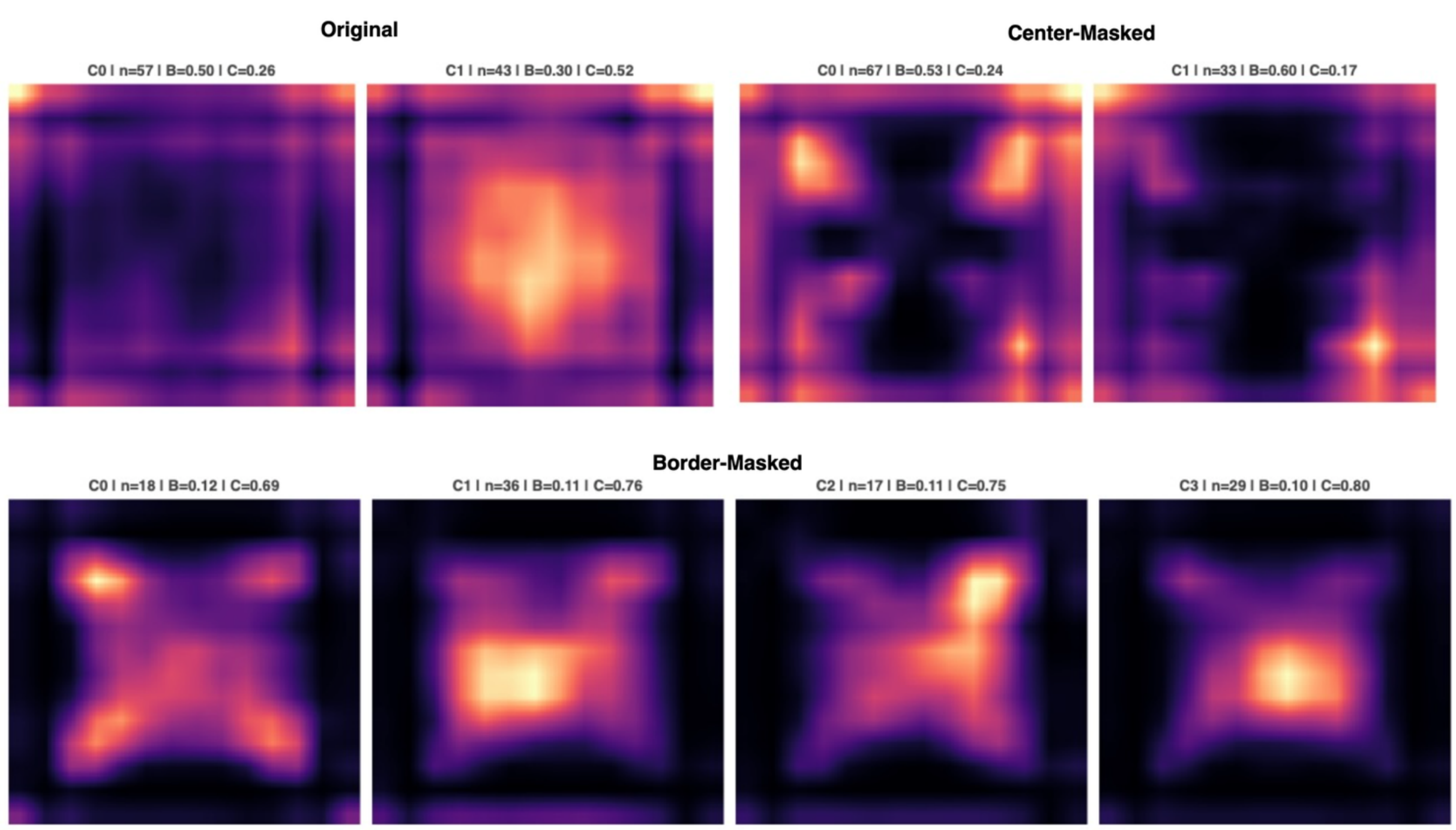

***Fig. 12. Masked-image ReMoDEx analysis showing region-dependent decision strategies under original, center-masked, and border-masked chest X-ray conditions.*** *The figure presents cluster prototypes obtained after applying Global Relevance Analysis to relevance maps generated under three input conditions. The* ***Original*** *condition shows the baseline relevance patterns without masking, the* ***Center-Masked*** *condition evaluates how the model behaves when central thoracic information is suppressed, and the* ***Border-Masked*** *condition evaluates how the model behaves when peripheral/border regions are removed. The shift in relevance patterns across these conditions indicates that the classifier's decision strategy is sensitive to both lung-centered and border-associated image regions.*

When the centre was masked, two clusters were found, both border/corner-oriented: centre relevance fell to 0.2383 and 0.1738 while border relevance rose to 0.5307 and 0.5972. At the condition level, Table 8 also shows a clear shift toward border/corner relevance, with the overall border/centre relevance ratio changing to 0.5520/0.2199. This condition produced the lowest fixed target score of 0.3880, the highest confidence drop of 0.4602, and an overall prediction-change rate of 0.55. The clustering statistics in Table 9 further show a two-cluster solution with silhouette 0.0920, Davies–Bouldin 2.8618, Calinski–Harabasz 10.3082, and mean entropy 0.9563. Both clusters became

COVID-dominant (true dominant labels Normal and Viral Pneumonia), and this condition produced the greatest disruption, with fixed target scores of 0.4203 and 0.3225 and prediction-change rates of 0.5224 and 0.6061 presented in Table 10. Further It shows in Table 10 that Cluster 0 was larger than Cluster 1 under centre masking, containing 67 samples compared with 33 samples, indicating that most masked-condition relevance maps converged toward a border/corner-oriented pattern. The higher prediction-change rate in Cluster 1, together with its lower fixed target score, suggests that the Viral Pneumonia-dominant group was more strongly disrupted by centre masking than the Normal-dominant group.

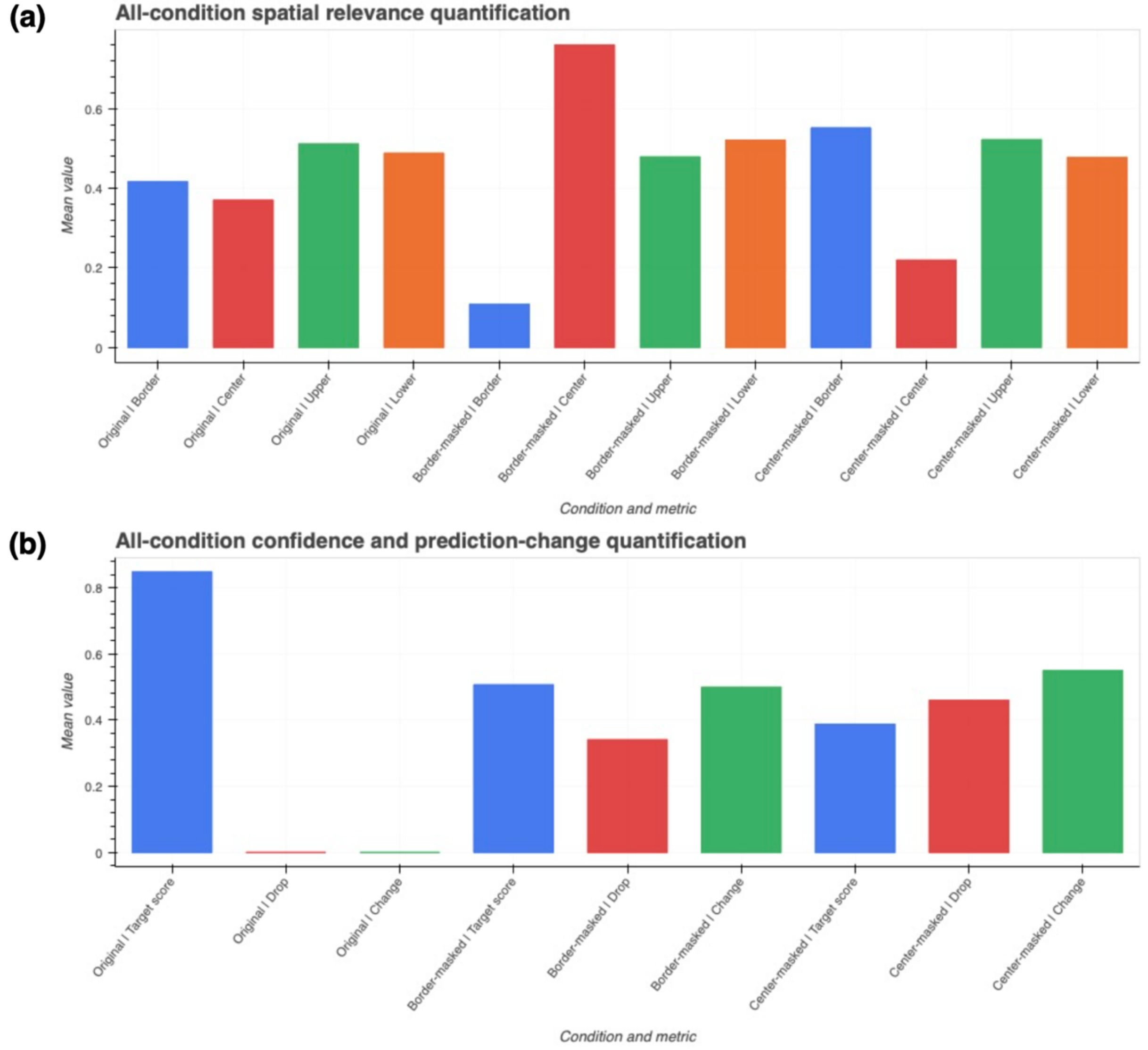

***Fig. 13. Quantitative assessment of spatial relevance and prediction changes across original, center-masked, and border-masked chest X-ray conditions.*** *The figure summarizes how masking different image regions affects the classifier's relevance distribution and prediction behaviour. Panel* ***(a)*** *shows the all-condition spatial relevance quantification, comparing border, center, upper, lower, and entropy-based relevance measures across original, center-masked, and border-masked conditions. Panel* ***(b)*** *shows the corresponding confidence and prediction-change quantification, including original confidence, current confidence after masking, fixed-target score, confidence drop, and prediction-change rate. Together, these results indicate that the model's decision behaviour changes when central thoracic or peripheral border regions are masked, supporting region-dependent decision sensitivity.*

As preseted in Fig. 13, masking altered both spatial relevance and prediction behaviour. Border masking moved relevance toward the central thoracic region (centre 0.76, border 0.11) while reducing the fixed target score to 0.51 and changing 50% of predictions. Centre masking moved relevance toward the periphery (border 0.55, centre 0.22) and caused the greatest loss of original target-class support, with the fixed target score falling to 0.39, the largest confidence drop (0.46), and the highest prediction-change rate (0.55). The low silhouette scores across conditions confirm that the clusters should be read as recurring relevance-pattern groups rather than distinct classes. The consistent shift to COVID predictions after masking indicates that spatial occlusion had a substantial effect on class-specific decision stability. Together with the multi-method clustering of Section 6.2, this validation provides causal support that the model's decisions depend on both central anatomical information and peripheral image cues.

# 7. Discussion

The central observation of this study is a clear gap between what conventional metrics report and what the model actually does. Judged on accuracy (86.27% test), AUC (0.9624), and balanced precision, recall, and F1-score, the VGG16 classifier would be accepted as reliable. Yet ReMoDEx shows that these correct predictions were not produced by a single, uniform decision strategy: across the assessed samples the framework consistently recovered two organisations of relevance—one concentrated in the central thoracic region and one dominated by image borders and corners. This directly instantiates, at the level of an individual trained model, the more general warnings raised in the literature: that deep networks readily exploit shortcuts rather than task-relevant signal (Geirhos et al., 2020), that such Clever Hans behaviour can hide behind strong benchmark numbers (Lapuschkin et al., 2019), and that COVID-19 radiograph classifiers in particular tend to select confounding, non-pathological cues over disease signal (DeGrave et al., 2021; Roberts et al., 2021). Zech et al. (2018) showed the same principle across hospitals, where models latched onto site-specific markers rather than pathology; the border/corner strategy surfaced here is the intra-dataset analogue of that effect. The value of ReMoDEx is that it makes this heterogeneity explicit and measurable instead of leaving it as an inference that only external testing can expose.

A distinctive strength of the present analysis is that the two-strategy structure emerged when each explainer was combined with the global module independently, rather than from any single method. The central-versus-peripheral split appeared under gradient-based Grad-CAM++ (Selvaraju et al., 2017; Chattopadhay et al., 2018), most sharply under perturbation-based Occlusion Sensitivity (Zeiler & Fergus, 2014) (centre 0.61/border 0.17 versus border 0.81/centre 0.05, the cleanest separation, as expected for a direct causal perturbation test), and in weaker form under path-integral-based Integrated Gradients (Sundararajan et al., 2017) and conservation-based Layer-wise Relevance Propagation (Bach et al., 2015). Because these families are grounded in different principles and fail in different ways, their agreement is far stronger evidence of a genuine model property than any one heatmap could provide. This speaks directly to a recurring criticism of post-hoc explainability—that individual explanations can be fragile, method-dependent, or even misleading, and that they may offer false reassurance in high-stakes settings (Rudin, 2019; Ghassemi et al., 2021). Prior chest X-ray studies have typically relied on a single explanation family, for example LRP in Bassi and Attux (2022) or an interpretability-and-uncertainty pipeline in Arias-Londoño and Godino-Llorente (2024); ReMoDEx complements these by turning the local-to-global relevance principle of Lapuschkin et al. (2019) into an explicit multi-method convergence test, so that a decision strategy is trusted only when independent explainers agree on it. This convergence-based reading is consistent with broader reviews that call for structured, quantitative use of relevance rather than isolated saliency inspection (Montavon et al., 2018; Samek et al., 2021; Arrieta et al., 2020; Tjoa & Guan, 2021).

The spatial content of the recovered strategies is itself interpretable in light of earlier work. The border/corner-dominant cluster is a candidate shortcut pattern of exactly the kind that other studies have traced to acquisition and dataset composition rather than pathology: source-specific and non-pathological cues (DeGrave et al., 2021), residual source bias that survives even after lung segmentation (Teixeira et al., 2021), background and non-target-region reliance that can be reduced by relevance-based optimisation (Bassi et al., 2024), and label-correlated artifacts such as chest drains that models latch onto (Jiménez-Sánchez et al., 2023). The co-existence of an anatomically plausible central-thoracic strategy with a peripheral one within the same well-performing model illustrates why relevance-based decision assessment is a necessary complement to accuracy: two subsets of correct predictions rested on qualitatively different evidence, and only one of them is clinically defensible. This aligns with the position that explanation analysis should evaluate whether attention lies in clinically meaningful regions, not merely whether predictions are correct (Reyes et al., 2020; Amann et al., 2020).

The masked-image validation converts these correlational patterns into causal evidence. Removing the periphery concentrated relevance centrally but destabilised the prediction, while removing the centre pushed relevance to the borders and caused the largest collapse in original-class support, with a majority of predictions flipping—most often to COVID; the single Viral Pneumonia case mirrored this, surviving border masking at reduced confidence but flipping to COVID once the centre was removed. The systematic drift toward one class when preferred evidence is occluded suggests the model falls back on a default rather than expressing uncertainty, echoing the reliability and fairness concerns raised when medical models depend on spurious or sensitive-attribute-correlated cues (Brown et al., 2023). It also complements the external-validation evidence of Bassi and Attux (2022) and the demonstration by Ong Ly et al. (2024) that hidden acquisition biases can inflate apparent performance by up to roughly twenty percent: where those works reveal fragility by changing the data distribution, ReMoDEx reveals the same fragility by perturbing the input regions the model relies on, without requiring a second dataset. For clinical translation, this is precisely the kind of failure mode that accuracy-first evaluation misses and that has been argued to stand between high benchmark performance and trustworthy deployment (Topol, 2019; Kelly et al., 2019).

Beyond the specific findings, the framework addresses a practical bottleneck in explainability for large-scale datasets. Local explanations are informative per prediction, but manually inspecting heatmaps for tens of thousands of images is impractical and time-consuming, and human verification cannot be applied to every sample. By passing each explainer's maps through a shared global module—built on standard, well-understood spectral clustering (Ng et al., 2002; von Luxburg, 2007)—ReMoDEx compresses an entire collection of relevance maps into a small number of decision-strategy prototypes that an analyst can read directly. In effect it recovers, from relevance alone, the kind of hidden subgroups that Oakden-Rayner et al. (2020) showed can drive clinically meaningful failures, but does so automatically and at dataset scale. Combined with the strong transfer-learning backbone used here (Simonyan & Zisserman, 2015; Deng et al., 2009; Shin et al., 2016) and the widely used COVID-19 Radiography Database (Chowdhury et al., 2020; Rahman et al., 2021), this makes chest X-ray classification a convenient case study, while the framework itself is dataset- and architecture-agnostic and applicable wherever per-sample human checking is infeasible (LeCun et al., 2015; Litjens et al., 2017).

Several limitations should temper the interpretation. The cluster-validity scores were modest (silhouette values below 0.10), so the identified groups are best understood as recurring relevance-pattern tendencies rather than sharply separated classes, and the border-dominant cluster is a candidate shortcut pattern rather than confirmed Clever Hans behaviour. The assessment was performed on a representative sample of predictions rather than the full test set, and on a single VGG16 architecture and one dataset. The masked-image validation used fixed geometric masks (outer 12% and central 60%), which approximate but do not perfectly isolate anatomical structures. ReMoDEx is a diagnostic and assessment tool, it characterises decision behaviour but does not by itself correct shortcut learning.

# 8. Conclusion

This study proposed ReMoDEx, a relevance-based local-to-global framework for assessing the decision behaviour of trained image classifiers. Rather than improving accuracy, ReMoDEx examines whether predictions are supported by task-relevant regions or by shortcut-sensitive structures, by drawing relevance evidence from four complementary local explainers—Grad-CAM++, Integrated Gradients, Occlusion Sensitivity, and Layer-wise Relevance Propagation—and consolidating them within a single global relevance-pattern analysis module. Applied to a VGG16 chest X-ray classifier with strong conventional performance (86.27% test accuracy, 0.9624 AUC), the framework revealed that correct predictions were produced by two recurring decision strategies, one central-thoracic and one border/corner-dominant, that were invisible to accuracy-based evaluation. Masked-image validation confirmed causal spatial sensitivity: border masking shifted relevance to the centre while reducing confidence, and centre masking shifted relevance to the periphery, produced the largest loss of original-class support, and flipped a majority of predictions. Because the global module condenses an entire collection of relevance maps into a few decision-strategy clusters, ReMoDEx is particularly suited to large-scale datasets, where inspecting every explanation heatmap by hand is impractical and time-consuming and human verification cannot be applied to each prediction; here chest X-ray classification serves only as a case study to demonstrate the framework. These findings establish ReMoDEx as a scalable relevance-based decision-assessment framework and an essential complement to accuracy-based evaluation, and demonstrate that it can surface candidate shortcut-learning and Clever Hans-type behaviour in clinically important classification systems. Future work will extend the framework to additional architectures, external datasets, anatomically informed masking, and relevance-guided retraining.

## Author contributions

Conceptualisation and design: Pathak A.K., Chaubey M. Supervision and validation: Pathak A.K., Chaubey M., Gupta M. Project administration: Chaubey M., Gupta M. Data collection and assembly: Pathak A.K. Data analysis and interpretation: Pathak A.K., Chaubey M. Visualisation: Pathak A.K. Writing — original draft: Pathak A.K., Chaubey M. Resources: Chaubey M., Gupta M. Writing — review and editing: Pathak A.K., Chaubey M., Gupta M.

## Funding

This research received no specific grant from any funding agency in the public, commercial, or not-for-profit sectors.

## Declaration of competing interest

The authors declare that they have no known competing financial interests or personal relationships that could have appeared to influence the work reported in this paper.

## Data availability

The dataset supporting the findings of this study is the publicly available COVID-19 Radiography Database, accessible through Kaggle. It contains chest X-ray images categorised into COVID-19 (3,616), Normal (10,192), Lung Opacity (6,012), and Viral Pneumonia (1,345). Class labels were derived from the original dataset directory structure and annotations provided by the maintainers. No private, institutional, or patient-identifiable data were used in this study.